\documentclass{new_tlp}

\usepackage[utf8]{inputenc}
\usepackage{amsmath,amssymb}
\usepackage{microtype}
\usepackage{url}

\usepackage{graphicx}

\usepackage{listings}
\providecommand{\sysfont}{\textit}

\newcommand{\clingo}{\sysfont{clingo}}

\newcommand{\plingo}{\sysfont{plingo}}
\newcommand{\Plingo}{\sysfont{Plingo}}

\providecommand{\Underscore}{\textunderscore}

\lstdefinelanguage{clingo}{%
  basicstyle=\ttfamily,%
  keywordstyle=[1]\bfseries,%
  keywordstyle=[2]\bfseries,%
  keywordstyle=[3]\bfseries,%
  showstringspaces=false,%
  literate={_}{\Underscore}1 {\%\%}{}0,%
  escapeinside={\#(}{\#)},%
  alsoletter={\#,\&},%
  keywords=[1]{not,from,import,def,if,else,elif,return,while,break,and,or,for,in,del,and,class,with,as,is,yield,async},%
  keywords=[2]{\#const,\#show,\#minimize,\#base,\#theory,\#count,\#external,\#program,\#script,\#end,\#heuristic,\#edge,\#project,\#show,\#sum},%
  keywords=[3]{&,&dom,&sum,&diff,&show},%
  morecomment=[l]{\#\ },%
  morecomment=[l]{\%\ },%
  morestring=[b]",%
  stringstyle={\itshape},%
  commentstyle={\color{darkgray}}%
}

\lstdefinelanguage{python}{%
  basicstyle=\ttfamily,%
  keywordstyle=[1]\bfseries,%
  showstringspaces=false,%
  literate={_}{\Underscore}{1},%
  escapeinside={\#(}{\#)},%
  alsoletter={\#,\&},%
  keywords=[1]{not,from,import,def,if,else,elif,return,while,break,and,or,for,in,del,and,class,with,as,is,yield,async},%
  morecomment=[l]{\#\ },%
  morestring=[b]",%
  stringstyle={\itshape},%
  commentstyle={\color{darkgray}}%
}

\newcommand{\lpmln}{\textit{Lpmln}}
\newcommand{\plog}{\textit{P-log}}
\newcommand{\problog}{\textit{ProbLog}}
\newcommand{\lpmlnl}{\textit{Lpmln}}
\renewcommand{\plingo}{\textit{Lpmln}\ensuremath{^{\mathit{\pm}}}}
\newcommand{\smproblog}{\textit{SMProblog}}

\newcommand{\sclingo}{\textit{clingo}}
\newcommand{\Splingo}{\textit{Plingo}}
\newcommand{\splingo}{\textit{plingo}}
\newcommand{\slpmln}{\textit{lpmln2asp}}

\newcommand{\splognaive}{\textit{plog-naive}}
\newcommand{\splogdco}{\textit{plog-dco}}
\newcommand{\sproblog}{\textit{problog}}
\newcommand{\saspmc}{\textit{aspmc}}

\newcommand{\unw}[1]{\ensuremath{{\overline{#1}}}}

\newcommand{\sm}[1]{\ensuremath{\mathit{SM}(#1)}}
\newcommand{\ssm}[1]{\ensuremath{\mathit{SSM}(#1)}}
\newcommand{\ssmp}[1]{\ensuremath{\mathit{SSM}^{\mathit{alt}}(#1)}}
\newcommand{\osm}[1]{\ensuremath{\mathit{OPT}^{{\pm}}(#1)}}
\newcommand{\os}[1]{\ensuremath{\mathit{OS}^{{\pm}}(#1)}}
\newcommand{\nos}[1]{\ensuremath{\mathit{NOS}^{{\pm}}(#1)}}

\newcommand{\tw}[1]{\ensuremath{\mathit{TW}(#1)}}

\newcommand{\osmplingo}[1]{\ensuremath{\mathit{OSM}^{\mathit{plingo}}(#1)}}

\newcommand{\weight}[2]{\ensuremath{W_{#1}(#2)}}
\newcommand{\weightp}[2]{\ensuremath{W_{#1}^{\mathit{alt}}(#2)}}
\newcommand{\weightpp}[2]{\ensuremath{W_{#1}^{{\pm}}(#2)}}
\newcommand{\weightplingo}[2]{\ensuremath{W_{#1}^{\mathit{plingo}}(#2)}}

\newcommand{\prob}[2]{\ensuremath{P_{#1}(#2)}}
\newcommand{\probbasic}[2]{\ensuremath{P^{\mathit{basic}}_{#1}(#2)}}
\newcommand{\probp}[2]{\ensuremath{P_{#1}^{\mathit{alt}}(#2)}}
\newcommand{\probpp}[2]{\ensuremath{P_{#1}^{{\pm}}(#2)}}
\newcommand{\probplingo}[2]{\ensuremath{P_{#1}^{\mathit{plingo}}(#2)}}

\newcommand{\soft}[1]{\ensuremath{#1^{\mathit{soft}}}}
\newcommand{\hard}[1]{\ensuremath{#1^{\mathit{hard}}}}
\newcommand{\weak}[1]{\ensuremath{#1^{\mathit{weak}}}}

\newcommand{\transtwo}[1]{\ensuremath{\mathit{standard}(#1)}}
\newcommand{\transtwozero}[0]{\ensuremath{\mathit{standard}}}
\newcommand{\transthree}[1]{\ensuremath{\mathit{alternative}(#1)}}
\newcommand{\transthreezero}[0]{\ensuremath{\mathit{alternative}}}
\newcommand{\transfour}[1]{\ensuremath{\mathit{negative}(#1)}}
\newcommand{\transtwoname}[1]{\ensuremath{\star}}
\newcommand{\transthreename}[1]{\ensuremath{\bullet}}

\newcommand{\totalcost}[2]{\ensuremath{\mathit{Cost}_{#1}(#2)}}
\newcommand{\cost}[3]{\ensuremath{\mathit{Cost}_{#1}(#2,#3)}}
\newcommand{\costplingo}[2]{\ensuremath{\mathit{Cost}_{#1}(#2,0)}}

\newtheorem{proposition}{Proposition}
\newtheorem{lemma}{Lemma}
\newcommand{\qed}{QED}
\newenvironment{proofof}[1]{\noindent {\bf Proof of #1.}}{\qed}

\newcommand{\problognormal}[1]{\ensuremath{{{#1}^{\mathit{normal}}}}}
\newcommand{\problognormalsub}[2]{\ensuremath{{{#1}_{#2}^{\mathit{normal}}}}}
\newcommand{\problogfacts}[1]{\ensuremath{{{#1}^{\mathit{probs}}}}}
\newcommand{\problogfactssub}[2]{\ensuremath{{{#1}_{#2}^{\mathit{probs}}}}}
\newcommand{\problogevidence}[1]{\ensuremath{{{#1}^{\mathit{evidence}}}}}
\newcommand{\problogatoms}[1]{\ensuremath{{{\mathit{choices}(#1)}}}}
\newcommand{\problogatomssub}[2]{\ensuremath{{{\mathit{choices}(#1_#2)}}}}
\newcommand{\problogtolpmlnzero}[0]{\ensuremath{{{\mathit{problog2lpmln}}}}}
\newcommand{\problogtolpmln}[1]{\ensuremath{{{\mathit{problog2lpmln}(#1)}}}}
\newcommand{\lpmlntoproblog}[1]{\ensuremath{{{\mathit{lpmln2problog}(#1)}}}}
\newcommand{\lpmlntoproblogzero}[0]{\ensuremath{{{\mathit{lpmln2problog}}}}}

\newcommand{\tr}[1]{\ensuremath{{\mathit{stratify}}(#1)}}
\newcommand{\posa}[1]{\ensuremath{#1'}}

\newcommand{\mybot}[0]{\ensuremath{\mathit{bot}}}

\newcommand{\atoms}[1]{\ensuremath{\mathit{atoms}(#1)}}
\newcommand{\softatoms}[1]{\ensuremath{\mathit{soft}(#1)}}

\newcommand{\pione}[0]{\Pi_1}
\newcommand{\pitwo}[0]{\Pi_2}
\newcommand{\pifinal}[0]{\problogtolpmln{\Pi}}
\newcommand{\pifinalnoev}[0]{\problogtolpmln{\Pi\setminus\problogevidence{\Pi}}}
\newcommand{\pifinalproblog}[0]{\Pi\setminus\problogevidence{\Pi}}
\newcommand{\ssum}[1]{\ensuremath{\mathit{sum}(#1)}}
\newcommand{\problogmodels}[1]{\ensuremath{\mathit{models}(#1)}}

\newtheorem{example}{Example}

\lstdefinelanguage{clingos}{%
  language=clingo,%
  basicstyle=\small\ttfamily%
}
\lstdefinelanguage{shells}{%
  basicstyle=\footnotesize\ttfamily,%
  numbers=none,%
}

\usepackage{footmisc}

\title{plingo: A system for probabilistic reasoning in Answer Set Programming}

\author[S.Hahn et al.]{%
  SUSANA HAHN \\
  {University of Potsdam, Germany and Potassco Solutions, Germany}
  \and
  TOMI JANHUNEN \\
  {Tampere University, Finland and Potassco Solutions Finland, Finland}
  \and
  ROLAND KAMINSKI
  \
  JAVIER ROMERO
  \
  NICOLAS R{\"U}HLING
  \
  TORSTEN SCHAUB
  \\ {University of Potsdam, Germany and Potassco Solutions, Germany}
}

\submitted{[n/a]}
\revised{[n/a]}
\accepted{[n/a]}

\begin{document}

\maketitle

\begin{abstract}
  We present \splingo, an extension of the ASP system \clingo\
  that incorporates various probabilistic reasoning modes.
  \Plingo\ is based on \plingo,
  a simple variant of the probabilistic language \lpmln,
  that follows a weighted scheme derived from Markov Logic.
  This choice is motivated by the fact that
  the main probabilistic reasoning modes can be mapped onto enumeration and optimization problems, and
  that \plingo\  may serve as a middle-ground formalism connecting to other probabilistic approaches.
  \Splingo\ offers three alternative frontends, for \lpmln, \plog, and \problog.
  These input languages and reasoning modes are implemented by means of \clingo's  multi-shot and
  theory solving capabilities.
  In this way, the core of \splingo\ is an implementation of \plingo\ in terms of modern ASP technology.
  On top of that, \splingo\ implements
  a new approximation technique based on a recent method for answer set enumeration in the order of optimality.
  Additionally, in this work we introduce a novel translation from \plingo\ to \problog.
  This leads to a new solving method in \splingo\ where
  the input program is translated and a \problog\ solver is executed.
  Our empirical evaluation shows that the different solving approaches of \splingo\ are complementary,
  and that \splingo\ performs similarly as other probabilistic reasoning systems.

  \medskip\noindent
  {\em Under consideration for publication in Theory and Practice of Logic Programming (TPLP)}
\end{abstract}



%
\section{Introduction}\label{sec:introduction}

Answer Set Programming (ASP; \citeN{lifschitz02a}) offers a rich knowledge representation language along
with a powerful solving technology.
While the paradigm has developed further, several probabilisitic extensions of ASP have been proposed,
among them \lpmln~\cite{leewan16a}, \problog~\cite{rakito07a}, and \mbox{\plog~\cite{bageru09a}}.

In this work, we present an extension of the ASP system~\clingo, called \splingo,
that features various probabilistic reasoning modes.
\Plingo\ is centered on \plingo, a simple variant of the probabilistic language \lpmln,
that is based upon a weighted scheme from Markov Logic~\cite{ricdom06a}.
\lpmln\ has already proven to be useful in several settings~\cite{leewan18a,ahlepasa19a} and
it serves us also as a middle-ground formalism connecting to other probabilistic modeling languages.
We rely on translations from \problog\ and \plog\ to \lpmln~\cite{leewan16a,leeyang17a}, respectively,
to capture these approaches as well.
In fact, \lpmln\ has already been implemented in the system \slpmln~\cite{letawa17a} by
mapping \lpmln-based reasoning into reasoning modes in \clingo\
(viz.\ optimization and enumeration of stable models).
As such, the core of \splingo\ can be seen as a re-implementation of \slpmln\ that is well integrated with \clingo\ by
using its multi-shot and theory reasoning functionalities.

In more detail,
the language \plingo\ constitutes
a subset of \lpmln\ restricting the form of weight rules
while being extended with ASP's regular weak constraints.
This restriction allows us to partition logic programs into two independent parts:
a \emph{hard} part generating optimal stable models, and
a \emph{soft} part determining the probabilities of these optimal stable models.
Arguably,
this separation yields a simpler semantics that leads in turn to an easier way of modeling probabilistic logic programs.
Nonetheless, it turns out that this variant is still general enough to capture full \lpmln.

The system \splingo\ implements the language \plingo\
within the input language of \clingo.
The idea is
to describe the hard part in terms of
normal rules and weak constraints at priority levels different from $0$,
and the soft part
via weak constraints at priority level~$0$. 
This fits well with the semantics of \clingo,
that considers higher priority levels more important. 
On top of this,
\splingo\ offers three alternative frontends for \lpmln, \plog, and \problog, respectively,
featuring dedicated language constructs that are in turn translated into the format described above. %
The frontends rely on
the translations from \problog\ and \plog\ to \lpmln\ from~\cite{leewan16a} and~\cite{leeyang17a},
respectively, and on our translation from \lpmln\ to \plingo.
For solving, the basic algorithm of \splingo\
follows the approach of \slpmln\ by reducing probabilistic reasoning to \clingo's
regular optimization and enumeration modes.
This is complemented by two additional solving methods.
The first
is an approximation algorithm that calculates probabilities using only the most probable $k$
stable models given an input parameter $k$.
This algorithm takes advantage of a new improved implementation
of the task of answer set enumeration in the order of optimality~(ASEO; \citeN{pajjan21a}).
The second method is based on a novel translation from \plingo\ to \problog, that we introduce in this paper.
The method translates the input program into a \problog\ program,
and then runs the system \sproblog~2.2~\cite{fibrreshguthjara15a} on that program.
Naturally, this approach benefits from the current and future developments of \problog\ solvers.
Interestingly, the solving techniques of \sproblog~2.2 are quite mature,
and they are different to the ones implemented in \splingo,
making the approach a good complement to the system.

We have empirically evaluated \splingo's performance by
comparing its different solving methods and by
contrasting them to original implementations of \lpmln, \problog\ and \plog.
The results
show that the different solving approaches of \splingo\ are indeed complementary,
and that \splingo\ performs at the same level as other probabilistic reasoning systems.

There are many other probabilistic extensions of logic programming.
For a recent review about them we refer the reader to~\cite{cozmau20} and the references therein.
Among the latest work most relevant to us,
from the knowledge representation perspective,
the credal semantics~\cite{cozmau20} and \smproblog~\cite{tokira21a}
can be seen as two different generalisations of
\problog\ to normal logic programs under stable models semantics.
From the implementation perspective,
\citeANP{eiheki21}~\citeyear{eiheki21}
present a method for algebraic answer set counting,
implemented in the system \saspmc,
and show promising results on its application to
probabilistic inference under stable model semantics.

The paper is organized as follows.
In Section 2, we review the necessary background material about \lpmln\ and \problog.
In Section 3, we define our language \plingo,
and introduce the translations from \lpmln\ to \plingo, and from \plingo\ to \problog.
We continue in Section 4 with the description of the language of the system \splingo.
There, we also present the frontends of \lpmln, \plog\ and \problog,
going through various examples.
Section 5 is devoted to the description of the implementation of \splingo,
and Section 6 describes the experimental evaluation.
We conclude in Section 7.
The proofs of the theoretical results are available in the Appendix.

This article extends the conference paper presented in~\cite{hajakarorusc22a}.
It includes many new examples to illustrate the formal definitions,
and the proofs of the formal results.
Most notably, this extended version introduces the translation from
\plingo\ to \problog, its implementation in \splingo,
and its experimental evaluation using \sproblog~2.2.
The comparison with \sproblog\ of~\cite{hajakarorusc22a}
had shown us that \sproblog\ could be much faster that \splingo\ and
the other ASP-based probabilistic systems.
This extension basically turns that handicap into an opportunity for \splingo.

%

\section{Background}\label{sec:background}

A \emph{logic program} is a set of propositional formulas.
A \emph{rule} is a propositional formula of the form
$H \leftarrow B$ where the head $H$ is a disjunction of literals and the body $B$ is either $\top$ or a conjunction of literals.
A rule is \emph{normal} if $H$ is an atom $a$, and it is a \emph{choice rule} if $H$ is a disjunction $a \vee \neg a$ for some atom $a$.
If $B$ is $\top$ we write simply $H$,
and if the rule is normal we call it a \emph{fact}.
%
%
%
%
%
We often identify a set of facts with the corresponding set of atoms.
A \emph{normal logic program} is a set of normal rules.
An interpretation is a set of propositional atoms.
An interpretation $X$ is a \emph{stable model} of a logic program $\Pi$
if it is a subset minimal model of the program that results
from replacing in $\Pi$ every maximal subformula that is not satisfied by $X$ by $\bot$~\cite{ferraris05a}.
The set of stable models of a logic program $\Pi$ is denoted by $\sm{\Pi}$.
%
%
A \emph{logic program with weak constraints}
is a set $\Pi_1 \cup \Pi_2$
where $\Pi_1$ is a logic program
and $\Pi_2$ is a set of weak constraints
of the form
\(
:\sim \, F[\mathit{w}, \mathit{l}]
\)
where $F$ is a formula,
$\mathit{w}$ is a real number weight, and
$\mathit{l}$ is a nonnegative integer.
The cost of a stable model $X$ of $\Pi_1$ at some nonnegative integer level $l$ is
the sum of the costs $w$ of the weak constraints
$:\sim \, F[\mathit{w}, \mathit{l}]$ from $\Pi_1$
whose formula $F$ is satisfied by $X$.
Given two stable models $X$ and $Y$ of $\Pi_1$,
$X$ is \emph{preferred} to $Y$ wrt.\ $\Pi_2$ if
there is some nonnegative integer $l$ such that
the cost of $X$ at $l$ is smaller than the cost of $Y$ at $l$,
and for all $l' > l$ the costs of $X$ and $Y$ at $l'$ are the same.
$X$ is an \emph{optimal model} of $\Pi_1 \cup \Pi_2$ if it is a stable model of $\Pi_1$
and there is no stable model $Y$ of $\Pi_1$ such that $Y$ is preferred to $X$ wrt.\ $\Pi_2$~\cite{bflnrp00a}.

We review the definition of \lpmln\ from~\cite{leewan16a}.
An \lpmln\ program $\Pi$ is a finite set of
weighted formulas $w : F$ where $F$ is a propositional formula and $w$ is either a real number
(in which case, the weighted formula is called soft)
or $\alpha$ for denoting the infinite weight (in which case, the weighted formula is called hard).
If $\Pi$ is an \lpmln\ program, by $\soft{\Pi}$ and $\hard{\Pi}$ we denote
the set of soft and hard formulas of $\Pi$, respectively.
For any \lpmln\ program $\Pi$ and any set $X$ of atoms,
$\unw{\Pi}$ denotes the set of (unweighted) formulas obtained from $\Pi$
by dropping the weights, and
$\Pi_X$ denotes the set of weighted formulas $w : F$ in $\Pi$ such that $X \models F$.
Given an \lpmln\ program $\Pi$, $\ssm{\Pi}$ denotes the set of \emph{soft stable models}
\(
\{X \mid X \textnormal{ is a stable model of } \unw{\Pi_X} \}.
\)
The \emph{total weight} of $\Pi$, written $\tw{\Pi}$, is defined as $\mathit{exp}(\sum_{w:F \in \Pi}w)$.

The \emph{weight} $\weight{\Pi}{X}$ of an interpretation
and its \emph{probability} $\prob{\Pi}{X}$
are defined, respectively, as
\begin{align*}
    \weight{\Pi}{X} & =
    \begin{cases}
        \tw{\Pi_X} & \textnormal{ if } X \in \ssm{\Pi} \\
        0          & \textnormal{ otherwise }
    \end{cases}
    \\
    \textnormal{ and }
    \\
    \prob{\Pi}{X}   & = \lim_{\alpha \rightarrow \infty}
    \frac{\weight{\Pi}{X}}{\sum_{Y \in \ssm{\Pi}} \weight{\Pi}{Y}} \, .
\end{align*}
An interpretation $X$ is called a \emph{probabilistic stable model} of $\Pi$ if $\prob{\Pi}{X} \neq 0$.

\par
\bigskip
\begin{example}
    Let $\Pi_1$ be the \lpmln\ program that consists of the following formulas:
    \[
        \begin{array}{c@{\hspace{25pt}}c}
            \alpha: a &
            1 : b
        \end{array}
    \]
    The soft stable models of $\Pi_1$ are $\emptyset$, $\{a\}$, $\{b\}$ and $\{a,b\}$.
    Their weights and probabilities are as follows:
    \begin{align*}
        \weight{\Pi_1}{\emptyset} & =0 & \weight{\Pi_1}{\{a\}} & =\mathit{exp}(\alpha) & \weight{\Pi_1}{\{b\}} & = \mathit{exp}(1) & \weight{\Pi_1}{\{a,b\}} & =\mathit{exp}(1+\alpha)
        \\
        \prob{\Pi_1}{\emptyset}   & =0 & \prob{\Pi_1}{\{a\}}   & \approx 0.269         & \prob{\Pi_1}{\{b\}}   & =0                & \prob{\Pi_1}{\{a,b\}}   & \approx 0.731
    \end{align*}
    To calculate the probabilities we first determine the denominator
    \begin{align*}
        \sum_{Y \in \ssm{\Pi}} \weight{\Pi}{Y} = 0 + \mathit{exp}(\alpha) + \mathit{exp}(1) + \mathit{exp}(1 + \alpha) \, .
    \end{align*}
    For the soft stable model $\{b\}$ we get
    \begin{align*}
        \prob{\Pi_1}{\{b\}} = \lim_{\alpha \rightarrow \infty}
        \frac{\mathit{exp}(1)}{\mathit{exp}(\alpha) + \mathit{exp}(1) + \mathit{exp}(1 + \alpha)} \, .
    \end{align*}
    If we apply $\lim_{\alpha \rightarrow \infty}$ we see that the denominator tends to infinity while
    the numerator is a constant. Therefore, the whole expression tends to $0$ as $\alpha$ approaches infinity.
    In the same way we can simplify the denominator by removing the $\mathit{exp}(1)$
    as it will always be dominated by the other terms containing $\mathit{exp}(\alpha)$.
    For the soft stable models $\{a\}$ and $\{a,b\}$ we then get the exact probabilities
    \(
    \prob{\Pi_1}{\{a\}} = 1 / (1+e)
    \)
    and
    \(
    \prob{\Pi_1}{\{a,b\}} = e / (1+e)
    \)
    whose approximate values we showed above.
    For the soft stable model $\{ \emptyset \}$ the probability is $0$ since the weight is $0$.

    Let $\Pi_2=\Pi_1 \cup \{\alpha: \neg a\}$.
    The soft stable models of $\Pi_2$ are the same as those of $\Pi_1$,
    but their weigths and probabilities are different:
    \begin{align*}
        \weight{\Pi_2}{\emptyset} & =\mathit{exp}(\alpha) & \weight{\Pi_2}{\{a\}} & =\mathit{exp}(\alpha) & \weight{\Pi_2}{\{b\}} & =\mathit{exp}(1+\alpha) & \weight{\Pi_2}{\{a,b\}} & =\mathit{exp}(1+\alpha)
        \\
        \prob{\Pi_2}{\emptyset}   & =0.134                & \prob{\Pi_2}{\{a\}}   & =0.134                & \prob{\Pi_2}{\{b\}}   & = 0.365                 & \prob{\Pi_2}{\{a,b\}}   & =0.365
    \end{align*}
    Let $\Pi_3$ and $\Pi_4$ replace the formula $1: b$ by
    the two formulas $\alpha : b \vee \neg b$ and $1 : \neg \neg b$ in $\Pi_1$ and $\Pi_2$, respectively,
    i.e., $\Pi_3 = \{ \alpha : a, \ \alpha : b \vee \neg b, \ 1 : \neg \neg b \}$ and
    $\Pi_4=\Pi_3 \cup \{\alpha: \neg a\}$.
    The soft stable models of $\Pi_3$ are the same as those of $\Pi_1$,
    their weights are the same as in $\Pi_1$ but incremented by $\alpha$,
    since all of them satisfy the new choice rule $\alpha : b \vee \neg b$,
    and their probabilities are the same as in $\Pi_1$.
    The same relation holds between the soft stable models of $\Pi_4$ and $\Pi_2$.
\end{example}

%
%
Besides the \emph{standard} definition,
we consider also an \emph{alternative} definition for \lpmlnl\ from \cite{leewan16a},
where soft stable models must satisfy all hard formulas of $\Pi$.
In this case, we have
\begin{align*}
    \ssmp{\Pi} = \{X \mid X \textnormal{ is a (standard) stable model of } \unw{\Pi_X}
    \textnormal{ that satisfies } \unw{\hard{\Pi}} \} \, ,
\end{align*}
while the \emph{weight} $\weightp{\Pi}{X}$ of an interpretation
and its \emph{probability} $\probp{\Pi}{X}$
are defined, respectively, as
\begin{align*}
    \weightp{\Pi}{X} & =
    \begin{cases}
        \tw{\soft{\Pi}_X} & \textnormal{ if } X \in \ssmp{\Pi} \\
        0                 & \textnormal{ otherwise }
    \end{cases}
    \\
    \textnormal{ and }
    \\
    \probp{\Pi}{X}   & = \frac{\weightp{\Pi}{X}}{\sum_{Y \in \ssmp{\Pi}} \weightp{\Pi}{Y}} \, .
\end{align*}
The set $\ssmp{\Pi}$ may be empty if there is no soft stable model that satisfies all hard formulas of $\Pi$,
in which case $\probp{\Pi}{X}$ is not defined.
On the other hand, if $\ssmp{\Pi}$ is not empty,
then for every interpretation $X$, the values of $\probp{\Pi}{X}$ and $\prob{\Pi}{X}$ are the same
(cf.~Proposition 2 from \cite{leewan16a}).

\begin{example}
    According to the alternative definition of \lpmln,
    the soft stable models of both $\Pi_1$ and $\Pi_3$ are $\{a\}$ and $\{a,b\}$.
    Their weights with respect to $\Pi_1$ are $\mathit{exp}(0)$ and $\mathit{exp}(1)$, respectively,
    and they are also $\mathit{exp}(0)$ and $\mathit{exp}(1)$ with respect to $\Pi_3$.
    The denominator is thus $1+e$ in both cases and
    therefore their probabilities are $1/(1+e)$ and $e/(1+e)$; the same as under the the standard definition.
    In turn, programs $\Pi_2$ and $\Pi_4$ have no soft stable models with the alternative semantics,
    and for this reason the probabilities of all of their interpretations are undefined.
\end{example}

In the next paragraphs,
we adapt the definition of \problog\ from~\cite{fibrreshguthjara15a} to our notation.
A \emph{basic} \problog\ program $\Pi$ consists of two parts:
a set \problognormal{\Pi} of normal rules,
and a set \problogfacts{\Pi} of probabilistic facts of the form $p :: a$
for some probability $p$ and some atom $a$.
Without loss of generality, we consider that $p$ is \emph{strictly} between $0$ and $1$.
By $\problogatoms{\Pi}$ we denote the set
$\{ a \mid p :: a \in \problogfacts{\Pi}\}$
of atoms occurring in the probabilistic facts of such a program.
We say that a basic \problog\ program is valid if it satisfies these conditions:
\begin{enumerate}
    \item
          The probabilistic atoms $a \in \problogatoms{\Pi}$ must not occur in any head $H$ of any rule $H \leftarrow B$ from \problognormal{\Pi}.
    \item
          For every set $X \subseteq \problogatoms{\Pi}$,
          the well-founded model~\cite{gerosc91a} of $\problognormal{\Pi}\cup X$ must be total.
\end{enumerate}
The second condition holds, in particular, if $\problognormal{\Pi}$ is positive or stratified~\cite{gerosc91a}.
Note also that, if the second condition holds, then the program $\problognormal{\Pi}\cup X$
has a unique stable model, that coincides with the true atoms of its well-founded model.
Following~\cite{fibrreshguthjara15a}, we consider only basic \problog\ programs that are valid.
\newpage
Given a basic \problog\ program $\Pi$, the probability $\prob{\Pi}{X}$ of an interpretation $X$
is defined as follows:
\begin{itemize}
    \item
          If $X$ is the (unique) stable model of $\problognormal{\Pi}\cup(X \cap \problogatoms{\Pi})$,
          then $\prob{\Pi}{X}$
          is the product of the products
          \begin{align*}
              \prod_{\substack{\phantom{a}p :: a \in \problogfacts{\Pi} \\ a \in X}}p
              \textnormal{ and }
              \prod_{\substack{\phantom{a}p :: a \in \problogfacts{\Pi} \\ a \notin X}}1 - p.
          \end{align*}
    \item
          Otherwise, $\prob{\Pi}{X}$ is $0$.
\end{itemize}

\begin{example}
    Let $\Pi_6$ be the \problog\ program that consists of the following elements:
    \[
        \begin{array}{c@{\hspace{25pt}}c}
            b \leftarrow \neg a &
            0.4 :: a
        \end{array}
    \]
    We have that
    $\problogfactssub{\Pi}{6}=\{ 0.4 :: a\}$, and
    $\problognormalsub{\Pi}{6}=\{ b \leftarrow \neg a\}$,
    $\problogatomssub{\Pi}{6}=\{a\}$.
    Program $\Pi_6$ is valid because
    it satisfies both condition 1,
    since the unique atom $a \in \problogatoms{\Pi_6}$ does not occur in the head of the unique rule of $\problognormalsub{\Pi}{6}$,
    and condition 2, since the program \problognormalsub{\Pi}{6} is stratified.
    The interpretations $\{a\}$ and $\{b\}$ have probability $0.4$ and $0.6$, respectively,
    and the others have probability $0$.
\end{example}

In~\cite{fibrreshguthjara15a}, the definition of an inference task may include the specification of some evidence.
Here, to simplify the presentation, it is convenient to include the evidence as part of a \problog\ program.
To do this, we represent this evidence by formulas of the form $\neg a$ or $\neg \neg a$ for some atom $a$,
that we call evidence literals.
We use evidence literals of the form $\neg \neg a$, instead of normal atoms of the form $a$,
to distinguish them clearly from normal facts, and to simplify our presentation later.
Then, we consider \emph{extended} \problog\ programs that
contain a set \problogevidence{\Pi} of evidence literals in addition to normal rules and probabilistic facts.
Both the notation and the definition of validity that we introduced above
carry over naturally to \problog\ programs of this extended form.
Just like before, we consider only valid extended \problog\ programs.
Next, let $\Pi$ be an extended \problog\ program.
If $X$ is an interpretation,
by \probbasic{\Pi}{X} we denote the probability
$\prob{\problognormal{\Pi}\cup\problogfacts{\Pi}}{X}$
of the corresponding basic \problog\ program.
Then, the probability of the evidence of $\Pi$ is
the sum of the basic probabilities \probbasic{\Pi}{X}
of the interpretations $X$ that satisfy all evidence literals in $\problogevidence{\Pi}$.
Finally, given an extended \problog\ program $\Pi$,
the probability $\prob{\Pi}{X}$ of an interpretation $X$
is:
\begin{itemize}
    \item undefined if the probability of the evidence of $\Pi$ is zero, otherwise
    \item it is $0$ if $X$ does not satisfy all evidence literals in $\problogevidence{\Pi}$,
          and otherwise
    \item
          it is the quotient between \probbasic{\Pi}{X}
          and the probability of the evidence of $\Pi$.
\end{itemize}
Basic \problog\ programs are a special case of extended \problog\ programs where
\problogevidence{\Pi} is empty.
From now on, we refer to extended \problog\ programs simply as \problog\ programs.

\begin{example}
    Let $\Pi_7=\Pi_6\cup\{\neg b\}$ be the \problog\ program that extends $\Pi_6$ by the evidence literal $\neg b$.
    It holds that \probbasic{\Pi_7}{X} is the same as \prob{\Pi_6}{X} for all interpretations $X$.
    Given this, the only interpretation of $\Pi_7$ with basic probability greater than $0$
    that satisfies the evidence literal $\neg b$ is $\{a\}$,
    whose basic probability is $0.4$.
    Hence, the probability of the evidence of $\Pi_7$ is $0.4$.
    Then, the probability of the interpretation $\{a\}$ is $1$,
    and it is $0$ for all other interpretations.
    If we replace the evidence literal $\neg b$ by $\neg \neg b$,
    then it is $\{b\}$ who has probability $1$, and the others have probability $0$.
    And if we add both $\neg b$ and $\neg \neg b$ at the same time,
    then the probabilities of all interpretations become undefined.
\end{example}
We close this section with the definition of the probability of a query atom $q$,
that is similar in both versions of \lpmln\ and in \problog:
%
it is undefined if the probability of the interpretations of the corresponding program is undefined,
which cannot happen with the standard definition of \lpmln,
and otherwise it is the sum of the probabilities of the interpretations that contain the query atom.
%

\section{The language \plingo\ }\label{sec:language}

In this section, we introduce the language \plingo\
and present translations from \lpmln\ to \plingo,
and from \plingo\ to \problog.
The former are used in the frontends of \lpmln, \problog\ and \plog,
combined with the translations from \problog\ and \plog\ to \lpmln\ from~\cite{leewan16a}
and~\cite{leeyang17a}, respectively.
The latter is used in the solving component of \splingo\,
to translate from \plingo\ to \problog\ and run a \sproblog\ solver.

The language \plingo\ is based on \lpmln\ under the alternative semantics.
The superscript $\pm$ in the name indicates that the new language both extends and restricts \lpmln.
The extension simply consists in adding weak constraints to the language.
This is a natural extension
that allows us to capture the whole \lpmln\ language
under both the alternative and the standard semantics.
On the other hand, the restriction limits the form of soft formulas
to \emph{soft integrity constraints} of the form $w : \neg F$
for some propositional formula $F$.
%
%
This is attractive because it allows us to provide
a definition of the semantics that is arguably very simple and intuitive.
Interestingly,
the translations from \problog\ and \plog~\cite{leewan16a,leeyang17a}
fall into this fragment of \lpmln.
Recall that in ASP, integrity constraints of the form $\neg F$ do not affect the generation of stable models,
but they can only eliminate some of the stable models generated by the rest of the program.
In \lpmln, soft integrity constraints parallel that role,
since they do not affect the generation of \emph{soft} stable models,
but they can only affect the probabilistic weights of
the \emph{soft} stable models generated by the rest of the program.
More precisely, it holds that
the soft stable models of an \lpmln\ program $\Pi$ remain the same
if we remove from $\Pi$ all its soft integrity constraints.
The reader can check that this is the case 
in our example programs $\Pi_3$ and $\Pi_4$ if we remove
their soft integrity constraint $1 : \neg \neg b$.
This observation leads us to the following proposition.
\begin{proposition}\label{prop:ssm}
    If $\Pi$ is an \lpmlnl\ program such that $\soft{\Pi}$ contains only soft integrity constraints,
    then $\ssmp{\Pi}=\sm{\unw{\hard{\Pi}}}$.
\end{proposition}
This allows us to leave aside the notion of soft stable models and
simply replace in $\weightp{\Pi}{X}$ and $\probp{\Pi}{X}$
the set $\ssmp{\Pi}$ by $\sm{\unw{\hard{\Pi}}}$.
From this perspective, an \lpmlnl\ program of this restricted form has two separated parts:
$\hard{\Pi}$, that generates stable models;
and $\soft{\Pi}$, that determines the weights of the stable models,
from which their probabilities can be calculated.

With these ideas, we can define the syntax and semantics of \plingo\ programs.
Formally, an \plingo\ program $\Pi$ is
a set of hard formulas, soft integrity constraints, and weak constraints,
denoted respectively by $\hard{\Pi}$, $\soft{\Pi}$ and $\weak{\Pi}$.
In what follows, we may identify
a hard formula or a set of them with their corresponding unweighted versions.
We say that $\Pi$ is \emph{normal} if $\hard{\Pi}$ is normal.
By $\osm{\Pi}$ we denote the optimal stable models of $\unw{\hard{\Pi}} \cup \weak{\Pi}$.
Then, the \emph{weight} and the \emph{probability} of an interpretation $X$,
written $\weightpp{\Pi}{X}$ and $\probpp{\Pi}{X}$, are defined as:
\begin{align*}
    \weightpp{\Pi}{X} & =
    \begin{cases}
        \tw{\soft{\Pi}_X} & \textnormal{ if } X \in \osm{\Pi} \\
        0                 & \textnormal{ otherwise }
    \end{cases}
    \\
    \textnormal{ and }
    \\
    \probpp{\Pi}{X}   & = \frac{\weightpp{\Pi}{X}}{\sum_{Y \in \osm{\Pi}} \weightpp{\Pi}{Y}} \, .
\end{align*}
Note that, as before,
$\osm{\Pi}$ may be empty, in which case $\probpp{\Pi}{X}$ is not defined.
Naturally, when \weak{\Pi} is empty the semantics coincide with the
alternative semantics for \lpmln.
In this case, \osm{\Pi} is equal to \sm{\unw{\hard{\Pi}}},
that by Proposition~\ref{prop:ssm} is equal to \ssmp{\Pi},
and both definitions are the same.

\begin{example}
    Programs $\Pi_1$ and $\Pi_2$ are not \plingo\ programs
    because they contain the soft formula $1 : b$.
    On the other hand,
    %
    $\Pi_3$ and $\Pi_4$ are \plingo\ programs,
    and they define the same probabilities as under the alternative semantics of \lpmln.
    Let us introduce the \plingo\ program $\Pi_5$,
    that replaces in $\Pi_3$ the formula $\alpha : a$ by
    the formulas $\alpha : a \vee \neg a$ and $:\sim \, a [-1,1]$,
    i.e., $\Pi_5 = \{
        \alpha : a \vee \neg a, \
        :\sim \, a [-1,1], \
        \alpha : b \vee \neg b, \
        1 : \neg \neg b \}$.
    The set \osm{\Pi_5} consists of the models $\{a\}$ and $\{a,b\}$,
    whose weights are $\mathit{exp}(0)$ and $\mathit{exp}(1)$, respectively,
    and whose probabilities are the same as in $\Pi_3$.

\end{example}

\subsection{From \lpmln\ to \plingo}\label{subsec:lpmlntoplingo}

We translate \lpmln\ programs to \plingo\ programs
following the idea of 
the translation \textit{lpmln2wc} from \cite{leeyang17a}.
%
An \lpmln\ program $\Pi$ under
the standard semantics is captured by the \plingo\ program
$\transtwo{\Pi}$
that contains:
\begin{itemize}
    \item
          the hard formulas
          $ \{  \alpha : F \vee \neg F \mid w:F \in \Pi \}$,
    \item
          the soft formulas
          $\{ w : \neg \neg F \mid w :F \in \Pi, w \neq \alpha \}$,
          and
    \item
          the weak constraints
          $\{ :\sim \, F [-1,1] \mid w : F \in \Pi, w = \alpha \}$.
\end{itemize}
The hard formulas generate the soft stable models of $\Pi$,
the weak constraints select those which satisfy most of the hard formulas of $\Pi$,
while the soft formulas attach the right weight to each of them,
without interfering in their generation.
The alternative semantics is captured by the translation
\transthree{\Pi} that contains:
\begin{itemize}
    \item
          the hard formulas $\{  \alpha : F \mid w:F \in \Pi, w = \alpha \} \cup
              \{  \alpha : F \vee \neg F \mid w:F \in \Pi, w \neq \alpha \}$,
    \item the same soft formulas as in \transtwo{\Pi}, and
    \item no weak constraints.
\end{itemize}
%
The first hard formulas enforce that the hard formulas of $\Pi$ must be satisfied,
while the latter are the same as in \transtwo{\Pi}, but only for the soft formulas of $\Pi$.
The weak constraints are not needed anymore.
\begin{proposition}\label{prop:probs_plingo}
    Let $\Pi$ be an \lpmlnl\ program.
    For every interpretation $X$, it holds that
    \begin{align*}
         & \prob{\Pi}{X} = \probpp{\transtwo{\Pi}}{X}
        \quad\textnormal{ and }                               \\
         & \probp{\Pi}{X} = \probpp{\transthree{\Pi}}{X} \, .
    \end{align*}
\end{proposition}

\begin{example}
    The \plingo\ programs $\Pi_3$, $\Pi_4$ and $\Pi_5$ of our examples are the result of
    applying the previous translations to the \lpmln\ programs $\Pi_1$ and $\Pi_2$.
    Namely,
    $\Pi_3$ is \transthree{\Pi_1},
    $\Pi_4$ is \transthree{\Pi_2}, and
    $\Pi_5$ is \transtwo{\Pi_1}.
    Accordingly, for all interpretations $X$, it holds that
    $\probp{\Pi_1}{X}=\probpp{\Pi_3}{X}$,
    $\probp{\Pi_2}{X}=\probpp{\Pi_4}{X}$, and
    $\prob{\Pi_1}{X}=\probpp{\Pi_5}{X}$.
    The program \transtwo{\Pi_2} is $\Pi_5\cup\{\alpha : \neg a \vee \neg \neg a, \ :\sim \, \neg a [-1,1]\}$.
\end{example}

As noted in~\cite{letawa17a},
these kinds of translations can be troublesome when
applied to logic programs with variables in the input language of \sclingo~\cite{cafageiakakrlemarisc20a}.
This is the case of the \lpmln\ frontend in \splingo,
where the rules at the input can be seen as
\emph{safe} implications $H \leftarrow B$
where $H$ is a disjunction and $B$ a conjunction of first-order atoms.
It is hard to see how to apply the previous translations
in such a way that the resulting soft formulas and weak constraints
belong to the input language of \sclingo,
since the result has to safisfy \sclingo's safety conditions.
For instance, if we try to apply the \transtwozero\ translation
to the hard rule \lstinline{a(X) :- b(X).}, 
a possible approach could generate the two weak constraints
\lstinline{:~ a(X). [-1,X]} and \lstinline{:~ not b(X). [-1,X]},
but the second of them is not safe and will not be accepted by \sclingo.
To overcome this problem,
we can use the \emph{negative} versions of the previous translations,
based on the translation \textit{lpmln2wc$^{pnt}$} from \cite{leeyang17a},
where the soft formulas for both translations are
\[\{ -w : \neg F \mid w :F \in \Pi, w \neq \alpha \},\]
and the weak constraints for the standard semantics are
\[\{ :\sim \, \neg F [1,1] \mid w : F \in \Pi, w = \alpha \}.\]
Observe that now $F$ always occurs under one negation.
In this case, when $F$ has the form $H \leftarrow B$,
the formulas $\neg F$ can be simply written as $\neg H \wedge B$, and
this formulation can be easily incorporated into \sclingo.
For instance,
\lstinline{a(X) :- b(X).}
is translated in this way to
\lstinline{:~ not a(X), b(X). [1,X]},
which is safe and accepted by \sclingo.
These negative versions are the result of applying
to \transtwo{\Pi} and \transthree{\Pi}
the translation of the following proposition,
and then simplifying the soft formulas of the form $-w : \neg \neg \neg F$
to $-w : \neg F$.
\begin{proposition}\label{prop:probs_plingo_neg}
    Given an \plingo\ program $\Pi$, let \transfour{\Pi} be the program 
    \begin{align*}
        \hard{\Pi} \cup
        \{ -w : \neg F \mid \, w : F \in \soft{\Pi} \} \cup
        \{ :\sim \, \neg F [-w,l]\mid \, :\sim \, F [w,l] \in \weak{\Pi}\} \, .
    \end{align*}
    For every interpretation $X$, it holds that $\probpp{\Pi}{X}$ and $\probpp{\transfour{\Pi}}{X}$ coincide.
\end{proposition}
This proposition is closely related to Corollary 1 from \cite{leeyang17a}.

\begin{example}
    The program \transfour{\Pi_5} is
    $\{
        \alpha : a \vee \neg a, \
        :\sim \, \neg a [1,1], \
        \alpha : b \vee \neg b, \
        -1 : \neg \neg \neg b \}$.
    Its last formula can be simplified to $-1 : \neg b$.
    The optimal stable models of this program are $\{a\}$ and $\{a,b\}$,
    their weights are $\mathit{exp}(-1)$ and $\mathit{exp}(0)$, respectively,
    and their probabilities are the same as in $\Pi_5$.
\end{example}

\subsection{From \problog\ to \plingo\ and back}\label{subsec:plingotoproblog}

\citeN{leewan16a} show how to translate \problog\ programs to \lpmln.
We obtain a translation from \problog\ to \plingo\
by combining that translation with our \transthreezero\ translation from \lpmln\ to \plingo.
Recall that we may identify
a hard formula or a set of them by their corresponding unweighted versions.
%
Let $\Pi$ be a \problog\ program, then the \plingo\ program \problogtolpmln{\Pi} is:\footnote{Note that
the soft formulas $\mathit{ln}(p/(1-p)) : \neg \neg a$
    could be alternatively represented by
    the pairs of soft formulas $\mathit{ln}(p) :: \neg \neg a$ and $\mathit{ln}(1-p) :: \neg a$.}
    \begin{align*}
        \problognormal{\Pi} \cup \problogevidence{\Pi} \cup
        \{ a \vee \neg a \mid p :: a \in \problogfacts{\Pi}\} \cup
        \{ \mathit{ln}(p/(1-p)) : \neg \neg a \mid p :: a \in \problogfacts{\Pi}\}.
    \end{align*}
    \begin{proposition}\label{prop:problogtoplingo}
        Let $\Pi$ be a \problog\ program.
        For every interpretation $X$, it holds that $\prob{\Pi}{X}$ and $\probpp{\problogtolpmln{\Pi}}{X}$ are the same.
    \end{proposition}

    \begin{example}
        Given our previous \problog\ program
        $\Pi_7=\{ b \leftarrow \neg a, \ \neg b, \ 0.4 :: a \}$,
        the \plingo\ program \problogtolpmln{\Pi_7}
        consists of the following formulas:
        \[\begin{array}{c@{\hspace{25pt}}c@{\hspace{25pt}}c@{\hspace{25pt}}c}
                b \leftarrow \neg a & \neg b & a \vee \neg a & -0.405 : \neg \neg a
            \end{array}\]
        where $-0.405$ is the result of $\mathit{ln}(0.4/(1-0.4))$.
        It holds that $\probpp{\problogtolpmln{\Pi_7}}{X}$ is $1$ when $X=\{a\}$,
        and it is $0$ otherwise, which is the same as \prob{\Pi_7}{X}.
    \end{example}

    In the reminder of this section,
    we present a translation in the other direction, from \plingo\ to \problog.
    The translation applies to non-disjunctive \plingo\ programs without weak constraints.
    At first sight, it may seem counterintuitive that such a translation is possible,
    since \problog\ is based on the well-founded semantics, and 
    the second condition for valid \problog\ programs severely restricts the form of their normal part.
    However, a closer look reveals that this restriction can be compensated
    by the other components of \problog's programs: probabilistic facts and evidence literals.
    As we will see,
    they can fulfill the role of choice rules and integrity constraints in ASP, respectively.
    Under this view, \problog\ programs resemble logic programs
    that follow the \emph{Generate, Define and Test} methodology~\cite{lifschitz02a}
    where probabilistic facts generate possible solutions,
    normal rules define additional predicates, and
    evidence literals filter the actual solutions.
    This relation makes the existence of a translation more intuitive.
    We make it precise in the next paragraphs.

    We present the translation for a \emph{normal} \plingo\ program $\Pi$
    without weak constraints,
    whose soft formulas have the form $w : \neg \neg a$ for some atom $a$.
    We assume that in $\Pi$ there are no different soft formulas
$w_1 : \neg \neg a$ and $w_2 : \neg \neg a$ for the same atom $a$.
    Using well-known translations,
    it is easy to extend the results to more general types of \plingo\ programs,
    as long as the complexity of deciding the satisfiability of the hard part remains in $\mathit{NP}$,
    and the programs contain no weak constraints.
    In fact, the implementation of this translation in our system \splingo\
    works for non-disjunctive \clingo\ programs~\cite{cafageiakakrlemarisc20a}.

    We modify $\Pi$ in 4 steps until we have a \plingo\ program
    that, in Step 5, we can easily turn into a \problog\ program
    %
    by inverting the translation \problogtolpmlnzero.
    We take as our running example the following \plingo\ program $\Pi_8$
    that is the result of applying to $\Pi_3$ the usual translation from choice rules to normal rules:
    \[ 
        \begin{array}{c@{\hspace{35pt}}c@{\hspace{35pt}}c@{\hspace{35pt}}c}
            a                             &
            b \leftarrow \neg \mathit{nb} &
            \mathit{nb} \leftarrow \neg b &
            1 : \neg \neg b
        \end{array}
    \] 
    The (optimal) stable models of $\Pi_8$ are $\{a,\mathit{nb}\}$ and $\{a,b\}$.
    Their probabilities are $0.269$ and $0.731$, respectively,
    just like those of $\{a\}$ and $\{a,b\}$ with respect to $\Pi_3$.

    \emph{Step 1.} We assume that the atom $\mybot$ does not occur in $\Pi$,
    and we add the literal $\neg \mybot$ to $\Pi$.
    In the end, this will be the unique evidence literal in the resulting \problog\ program.
    Integrity constraints $\bot \leftarrow B$ are not allowed in \problog,
    but once we have the evidence literal $\neg \mybot$,
    we can represent them simply by $\mybot \leftarrow B$.
    This shows how evidence literals fulfill the role of integrity constraints.

    \emph{Step 2.} For every atom $a$ occurring in $\Pi$, we add the following rules
    introducing a new atom \posa{a} that works as a copy of $a$:
    \begin{align}\label{eq:copy}
        \begin{array}{c@{\hspace{25pt}}c@{\hspace{25pt}}c}
            \posa{a} \vee \neg \posa{a}              &
            \mybot \leftarrow a \wedge \neg \posa{a} &
            \mybot \leftarrow \neg a \wedge  \posa{a}
        \end{array}
    \end{align}
    The choice rule selects a truth value for $\posa{a}$, while
    the other rules act as integrity constraints that enforce
    the truth values of $a$ and \posa{a} to be the same.
    After adding $\neg \mybot$ and these rules to $\Pi$,
    the resulting \plingo\ program has the same stable models as before,
    but for every atom $a$ in a stable model we also have its copy $\posa{a}$.
    Apart from this, the probabilities of the stable models remain the same.
    In our example, we add to $\Pi_8$ the literal $\neg \mybot$,
    as well as the formulas~\eqref{eq:copy} for the three atoms $a$, $b$ and $\mathit{nb}$
    occurring in $\Pi_8$.
    The stable models are now
$\{a,\posa{a},\mathit{nb},\posa{\mathit{nb}}\}$ and
$\{a,\posa{a},\mathit{b},\posa{\mathit{b}}\}$,
    and their probabilities are, as before, $0.269$ and $0.731$.

    \emph{Step 3.} To be able to satisfy condition 2 of valid \problog\ programs,
    we turn the set of normal rules of the original program into a set of stratified rules,
    by replacing every negative literal $\neg a$ occurring in them by $\neg \posa{a}$.
    That is, we replace every normal rule $r$ in the original program of the form:
    \[
        a_0 \leftarrow a_1 \wedge \ldots \wedge a_m \wedge \neg {a_{m+1}}, \ldots, \neg {a_n}
    \]
    by the normal rule \tr{r}:
    \[
        a_0 \leftarrow a_1 \wedge \ldots \wedge a_m \wedge \neg {\posa{a}_{m+1}}, \ldots, \neg {\posa{a}_n}.
    \]
    This replacement does not affect the stable models of the program,
    given that $a$ and $\posa{a}$ are equivalent and
    the replacement only happens in negative literals.
    If we also replaced the atoms $a_1$ to $a_m$ by $\posa{a}_1$ to $\posa{a}_m$,
    then the resulting program would represent the supported models of the original program instead of the stable ones.
    It is easy to see that the resulting set of rules is stratified
    because the atoms $\posa{a}$ occurring in the negative literals do not occur in any head.
    In our example, the three normal rules of $\Pi_8$ are replaced by these ones:
    \[
        \begin{array}{c@{\hspace{35pt}}c@{\hspace{35pt}}c} 
            a                                    &
            b \leftarrow \neg \posa{\mathit{nb}} &
            \mathit{nb} \leftarrow \neg \posa{b}
        \end{array}
    \]

    \emph{Step 4.} To invert the translation \problogtolpmlnzero,
    we have to translate
    every pair of a choice rule
    and a soft formula
    of our current \plingo\ program
    into a probabilistic fact.
    , for doing this, we need such a pairing between choice rules and soft formulas.
    We achieve it as follows.
    We replace every soft formula of the form $w : \neg \neg a$ by $w : \neg \neg \posa{a}$.
    This does not change the probabilities of the stable models, since $a$ and $\posa{a}$ are equivalent.
    Additionally, for the atoms $a$ not occurring in the soft formulas,
    we add the trivial formula $0 : \neg \neg \posa{a}$.
    Clearly, this does not affect the probabilities of the stable models.
    At this point, for every choice rule $\posa{a} \vee \neg \posa{a}$ in our \plingo\ program\
    there is one soft formula $w : \neg \neg \posa{a}$, and vice versa.
    In our example, the previous soft formula $1 : \neg \neg b$ is replaced by these ones:
    \[
        \begin{array}{c@{\hspace{35pt}}c@{\hspace{35pt}}c} 
            0 : \neg \neg \posa{a} &
            1 : \neg \neg \posa{b} &
            0 : \neg \neg \posa{\mathit{nb}}
        \end{array}
    \]

    \emph{Step 5.} To finalize, we just have to invert the translation \problogtolpmlnzero.
    We do it by replacing every pair of
    a choice rule $\posa{a} \vee \neg \posa{a}$ and a soft formula $w : \neg \neg \posa{a}$
    by the probabilistic fact $e^w / (e^w + 1) :: \posa{a}$.
    The function from $w$ to $e^w / (e^w + 1)$ to calculate the probabilities
    is simply the inverse of the function from $p$ to $\mathit{ln}(p/(1-p))$ to calculate the weights in
    \problogtolpmlnzero. 
    In particular, for $w=0$ it leads to the probability $0.5$.
    Note how probabilistic facts fulfill the role of the choice rules,
    and at the same time they stand for the soft formulas.
    In our example, the choice rules from~\eqref{eq:copy} and the previous soft formulas
    are replaced by the following probabilistics facts:
    \[
        \begin{array}{c@{\hspace{35pt}}c@{\hspace{35pt}}c} 
            0.5 :: \posa{a}               &
            e/(e+1) :: \neg \neg \posa{b} &
            0.5 :: \posa{\mathit{nb}}
        \end{array}
    \]
    In the resulting \problog\ program, the interpretations
$\{a,\posa{a},\mathit{nb},\posa{\mathit{nb}}\}$ and
$\{a,\posa{a},\mathit{b},\posa{\mathit{b}}\}$
    have probabilities $0.269$ and $0.731$, respectively,
    and the other interpretations have probability $0$.
    This is the same as in $\Pi_8$,
    once we eliminate the additional atoms $\posa{a}$, $\posa{b}$, and $\posa{\mathit{nb}}$.

    Finally, we can put all the steps of the translation together.
    Let $\Pi$ be our original program,
    let \atoms{\Pi} denote the set of atoms of $\Pi$,
    and \softatoms{\Pi} denote the set of atoms
$\{a \mid w : \neg \neg a \in \soft{\Pi}\}$
    occurring in the soft formulas of $\Pi$.
    Then, the \problog\ program \lpmlntoproblog{\Pi} consists of:
    \begin{itemize}
        \item
              the normal rules
              \begin{align*}
                  \{\mybot \leftarrow a \wedge \neg \posa{a} & \mid a \in \atoms{\Pi}\} \ \cup \\
                  \{\mybot \leftarrow \neg a \wedge \posa{a} & \mid a \in \atoms{\Pi}\} \ \cup \\
                  \{ \tr{r}                                  & \mid r \in \hard{\Pi}\},
              \end{align*}
        \item
              the probabilistic facts
              \begin{align*}
                  \{ e^w / (e^w + 1) :: \posa{a} \mid w : \neg \neg a \in \soft{\Pi} \} & \ \cup \\
                  \{ 0.5 :: \posa{a} \mid a \in \atoms{\Pi}\setminus\softatoms{\Pi}\}   & ,
              \end{align*}
        \item
              and the evidence literals \[\{\neg \mybot\}.\]
    \end{itemize}
    Observe that this is a valid \problog\ program.
    Condition 1 of validity is satisfied because the atoms $\posa{a}$ occurring in the probabilistic
facts do not occur in any head of the normal rules, and
condition 2 is satisfied because the normal rules are stratified.
The following proposition states the correctness of the translation.
\begin{proposition}\label{prop:plingotoproblog}
    Let $\Pi$ be a \emph{normal} \plingo\ program $\Pi$ without weak constraints,
    whose soft formulas have the form $w : \neg \neg a$ for some atom $a$
    that has at most one formula of that form.
    For every interpretation $X$ disjoint from $\{\posa{a}\mid a \in \atoms{\Pi}\}$
    it holds that $\probpp{\Pi}{X}$ and $\prob{\lpmlntoproblog{\Pi}}{X\cup\{\posa{a}\mid a \in X\}}$ are the same.
    %
\end{proposition}

\section{The language of \splingo\ and its frontends }\label{sec:plingo}

In this section we first describe the core language of \splingo,
that basically re-interprets the language of \clingo\ in terms of the semantics of \plingo.
After that,
we illustrate the frontends of \splingo\ with examples,
showing in each case what is the result of
the translation to the core language of \splingo.

The main idea of \splingo\ is to keep the input language of \clingo,
and re-interpret weak constraints at priority level $0$ as soft integrity constraints.
As explained above, these constraints are not considered to determine the optimal stable models,
but instead are used to determine the weights of those models,
from which their probabilities are calculated.
%
%
For programs in the input language of \splingo\ (or of \clingo, that is the same)
we can in fact provide a general definition that relies on the
definitions used for \clingo\ \cite{cafageiakakrlemarisc20a},
and that therefore covers its whole language.
We define a \splingo\ program $\Pi$ as a logic program in the language of \clingo,
and we let \osmplingo{\Pi} denote the optimal stable models of $\Pi$ without considering weak constraints at level $0$,
and \costplingo{\Pi}{X} denote the cost of the interpretation $X$ at priority level $0$,
according to the definitions of \cite{cafageiakakrlemarisc20a}.
Then, the \emph{weight} and the \emph{probability} of an interpretation $X$,
written $\weightplingo{\Pi}{X}$ and $\probplingo{\Pi}{X}$, respectively,
are analogous to $\weightp{\Pi}{X}$ and $\probp{\Pi}{X}$,
but replacing the set $\ssmp{\Pi}$ by $\osmplingo{\Pi}$:
%
\begin{align*}
    \weightplingo{\Pi}{X} & =
    \begin{cases}
        \mathit{exp}(\cost{\Pi}{X}{0}) & \textnormal{ if } X \in \osmplingo{\Pi} \\
        0                              & \textnormal{ otherwise }
    \end{cases}
    \\
    \textnormal{ and }
    \\
    \probplingo{\Pi}{X}   & = \frac{\weightplingo{\Pi}{X}}{\sum_{Y \in \osmplingo{\Pi}} \weightplingo{\Pi}{Y}} & \, .
\end{align*}

%

\subsection{The frontend of \lpmln}
Listing~\ref{lst:birds} shows the birds example from \cite{leewan16a}
using the frontend of \lpmln.
To start with, there is some general knowledge about birds:
both resident birds and migratory birds are birds,
and a bird cannot be both resident and migratory.
This is represented by the hard rules in Lines~\ref{lst:birds:encoding:begin}\nobreakdash-\ref{lst:birds:encoding:end},
that are written as common \sclingo\ rules.
Additionally,
from one source of information we have the fact that $\texttt{jo}$ is a resident bird,
while from another we have that $\texttt{jo}$ is a migratory bird.
For some reason, we hold the first source to be more trustworthy than the second.
This information is represented by the soft rules in Lines~\ref{lst:birds:weight1} and~\ref{lst:birds:weight2},
where the weights are expressed by the (integer) arguments of their \texttt{\&weight/1} atoms in the body.
The first soft rule corresponds to the weighted formula $2 : resident(jo)$,
and the second to $1 : migratory(jo)$.
Under both the standard and the alternative semantics,
this program has three probabilistic stable models:
$\{\}$, $\{{resident(jo)}, {bird(jo)}\}$,
and $\{{migratory(jo)}, {bird(jo)}\}$,
whose probabilities are $0.09$, $0.67$, and $0.24$, respectively.
They can be computed by \splingo,
running the command
\lstinline{plingo --mode=lpmln birds.plp}
for the standard semantics,
and using the option \lstinline{--mode=lpmln-alt}
for the alternative semantics.

\Splingo\ translates \lpmln\ programs using the
\emph{negative} versions of the \transtwozero{} and \transthreezero{} translations
from Section~\ref{subsec:lpmlntoplingo}.
Considering first the alternative semantics,
the hard rules remain the same,
while the soft ones are translated as shown in Listing~\ref{lst:birds_plingo}.
According to the negative version of the \transthreezero{} translation,
the soft formula $2 : resident(jo)$ becomes
the hard formula $\alpha : resident(jo) \vee \neg resident(jo)$ and
the soft formula $-2 : \neg resident(jo)$.
In \splingo, the first is written as the choice rule in Line~1,
and the second as the weak constraint at level $0$ of Line~2.
The translation of the other soft fact is similar.
Considering now the standard semantics,
the first rule of Listing~\ref{lst:birds}
becomes the choice rule
\lstinline|{bird(X)} :- resident(X)|
together with the weak constraint
\lstinline{:~ not bird(X), resident(X).[-1@1,X]}.
The second rule is translated similarly. 
The third one becomes simply
\lstinline{:~ resident(X), migratory(X).} \lstinline{[-1@1,X]},
since the additional choice rule is a tautology and can be skipped.
Observe that both weak constraints
use the variable \texttt{X}
in the expression \texttt{[-1@1,X]}.
This ensures that stable models obtain a weight of \texttt{-1}
for every ground instantiation of the corresponding body that they satisfy.


\begin{table}
    \begin{minipage}[t]{0.45\linewidth}
\begin{lstlisting}[caption={\lpmln\ birds example \newline (\texttt{birds.plp}).},language=clingo,label={lst:birds}]
bird(X) :- resident(X).%%#(\label{lst:birds:encoding:begin}#)
bird(X) :- migratory(X).%%#(\label{lst:birds:encoding:middle}#)
:- resident(X), migratory(X).%%#(\label{lst:birds:encoding:end}#)
resident(jo) :- &weight(2).%%#(\label{lst:birds:weight1}#)
migratory(jo) :- &weight(1).%%#(\label{lst:birds:weight2}#)
\end{lstlisting}
    \end{minipage}%
    \begin{minipage}[t]{0.1\linewidth}
        \phantom{A}
    \end{minipage}%
    \begin{minipage}[t]{0.45\linewidth}
\begin{lstlisting}[language=clingo,label={lst:birds_plingo},caption={Translation of the birds \newline example.},firstline=9,lastline=13]
{ resident(jo) }.                      %%#(\label{lst:birds:tr:6}#)
:~ not resident(jo). [-2@0]          %%#(\label{lst:birds:tr:7}#)

{ migratory(jo) }.                     %%#(\label{lst:birds:tr:8}#)
:~ not migratory(jo). [-1@0]         %%#(\label{lst:birds:tr:9}#)
\end{lstlisting}
    \end{minipage}%
\end{table}

\subsection{The frontend of \problog}

We illustrate the frontend of \problog\ with an example 
where we toss two biased coins whose probability of turning up heads is $0.6$.
We would like to know what is the probability of the first coin turning up heads,
given some evidence against the case that both coins turn up heads.
The representation in \splingo\ is shown in Listing~\ref{lst:coins}.
The first rule represents the toss of the coins.
Its ground instantiation leads to two probabilistic facts, one for each coin,
whose associated probabilities are specified by the \texttt{\&problog/1} atom in the body.
The argument of \texttt{\&problog/1} atoms is a string that contains either a float number or an expression, e.g., $``\texttt{3/5}"$.
Since the argument is a probability, the string must either contain or evaluate to a real number between $0$ and $1$.
The next line poses the query about the probability of the first coin turning up heads,
using the theory atom \texttt{\&query/1}, whose unique argument is an atom.
Finally, Lines~\ref{lst:coins:twoheads} and \ref{lst:coins:evidence} add the available evidence,
using the theory atom \texttt{\&evidence/2},
whose arguments are an atom and a truth value (\texttt{true} or \texttt{false}).
%
%
In \problog, the probabilistic facts alone lead to four possible worlds:
$\{\}$ with probability $0.4*0.4=0.16$,
$\{\texttt{heads(1)}\}$ and $\{\texttt{heads(2)}\}$ with probability $0.6*0.4=0.24$ each,
and $\{\texttt{heads(1)}, \texttt{heads(2)}\}$ with probability $0.6*0.6=0.36$.
The last possible world is eliminated by the evidence, and
we are left with three possible worlds.
Then, the probability of $\texttt{heads(1)}$ is the result of dividing
the probability of $\{\texttt{heads(1)}\}$ by the sum of the probabilities of the three possible worlds,
i.e., $\frac{0.24}{0.16+0.24+0.24}=0.375$.
This is the result that we obtain 
running the command \lstinline{plingo --mode=problog coins.plp}.

\Splingo\ translates \problog\ programs using the translation
\problogtolpmlnzero\ from Section~\ref{subsec:plingotoproblog}.
%
The result in this case is shown in Listing~\ref{lst:coins_plingo}.
In the propositional case,
the probabilistic \problog\ fact $0.6 :: heads(1)$
is translated to the weighted fact $w: heads(1)$,
where $w=ln(0.6/(1-0.6))\approx 0.40546$ \footnote{%
With this representation,
the weights do not stand for probabilities,
but for the logarithm of the probabilities.
Then, the cost of a stable model at level $0$
represents the sum of the logarithms of the relevant probabilities
and, by exponentiating that value, the probabilistic weight of a stable model
becomes the product of the corresponding probabilities.\label{footnote:log}
}
that in \plingo\ becomes
the hard formula $\alpha : heads(1) \vee \neg heads(1)$
together with the soft integrity constraint $w : \neg \neg heads(1)$.
The translation for the other probabilistic fact is similar.
In \splingo, for \texttt{C=1..2},
the hard formula is written as the choice rule of Line~\ref{lst:coins_plingo:prob1},
and the soft one is written as a weak constraint at level $0$ in the next line,
after simplifying away the double negation,
where \texttt{@f(X)} is an external function that returns the natural logarithm of \texttt{X/(1-X)}.
%
%
Going back to the original program,
the \texttt{\&query/1} atom is stored by the system
to determine what reasoning task to perform,
the normal rule in Line~\ref{lst:coins:twoheads} is kept intact, and
the \texttt{\&evidence/1} atom is translated to the integrity constraint
of Line~\ref{lst:coins_plingo:evidence},
that excludes the possibility of both coins turning up heads.

\vspace{3mm}
\begin{lstlisting}[language=clingo,label={lst:coins},caption={\problog\ tossing coins example (\texttt{coins.plp}).}]
heads(C) :- &problog("0.6"), C=1..2.%%#(\label{lst:coins:prob}#)
&query(heads(1)).%%#(\label{lst:coins:query}#)
two_heads :- heads(1), heads(2).%%#(\label{lst:coins:twoheads}#)
&evidence(two_heads, false).%%#(\label{lst:coins:evidence}#)
\end{lstlisting}

\begin{lstlisting}[language=clingo,label={lst:coins_plingo},caption={Translation of the coins example.}]
{ heads(C) }  :- C=1..2.%%#(\label{lst:coins_plingo:prob1}#)
:~ heads(C), C=1..2. [@f("0.6")@0,C]%%#(\label{lst:coins_plingo:prob2}#)
two_heads :- heads(1), heads(2).%%#(\label{lst:coins_plingo:twoheads}#)
:- two_heads.%%#(\label{lst:coins_plingo:evidence}#)
\end{lstlisting}


\subsection{The frontend of \plog}
\label{subsec:plog}

We illustrate the frontend of \plog\ with a simplified version of the dice example from~\cite{bageru09a},
where there are two dice of six faces.
The first dice is fair, while the second one is biased to roll $6$ half of the times.
We roll both dice, and observe that the first rolls a $1$.
We would like to know what is the probability of the second dice rolling another $1$.
The representation in \splingo\ using the \plog\ frontend is shown in Listing~\ref{lst:dice}.
%
Given that the original language \plog\ is sorted,
a representation in that language would contain the sorts
$\mathit{dice}=\{\texttt{d1},\texttt{d2}\}$ and
$\mathit{score}=\{\texttt{1}, \ldots, \texttt{6}\}$,
and the attribute $\mathit{roll}: \mathit{dice} \to \mathit{score}$.
In \splingo\ there are no attributes, and the sorts are represented by normal atoms,
like in the first two lines of Listing~\ref{lst:dice}.
Then, for example,
to assert that the result of rolling dice \texttt{d2} is \texttt{6},
in \plog\ one would write an assignment \texttt{roll(d2)=6}
stating that the attribute \texttt{roll(d2)} has the value \texttt{6},
while in \splingo\ one would use a normal atom of the form \texttt{roll(d2,6)}.
Going back to the encoding,
Line~\ref{lst:dice:random} contains a \emph{random selection rule}
that describes the experiments of rolling every dice \texttt{D}.
Each of these experiments selects at random one of the scores of the dice,
unless this value is fixed by a deliberate action of the form \texttt{\&do(A)},
that does not occur in our example.
Line~\ref{lst:dice:pr2} contains a \emph{probabilistic atom}
stating that the probability of dice \texttt{d2} rolling a \texttt{6} is $1/2$.
By the \emph{principle of indifference},
embodied in the semantics of \plog,
the probability of each of the $5$ other faces of \texttt{d2} is $(1-1/2)/5=0.1$,
while the probability of each face of \texttt{d1} is $1/6$.
Line~\ref{lst:dice:obs} represents the observation of the first dice rolling a $1$,
and the last line states the query about the probability of
the second dice rolling another $1$.
Running the command
\lstinline{plingo --mode=plog dice.plp},
we obtain that this probability is, as expected, $0.1$.
If we replace the query by \texttt{\&query(roll(d1,1))},
then we obtain a probability of $1$, and not of $1/6$,
because the observation in Line~\ref{lst:dice:obs} is
only consistent with the first dice rolling a $1$.
%
\begin{lstlisting}[float,language=clingo,label={lst:dice},caption={\plog\ dice example (\texttt{dice.plp}).}]
dice(d1;d2).%%#(\label{lst:dice:sorts:begin}#)
score(1..6).%%#(\label{lst:dice:sorts:end}#)
&random { roll(D,X) : score(X) } :- dice(D).%%#(\label{lst:dice:random}#)
&pr { roll(d2,6) } = "1/2".%%#(\label{lst:dice:pr2}#)
&obs{ roll(d1,1) } = true.%%#(\label{lst:dice:obs}#)
&query(roll(d2,1)).%%#(\label{lst:dice:query}#)
\end{lstlisting}
%
\begin{lstlisting}[float,language=clingo,label={lst:dice_meta},caption={Translation of the dice example.}]
random(roll(D),(roll(D),X)) :- score(X); dice(D).%%#(\label{lst:dice_meta:random}#)
pr(roll(d2),(roll(d2),6),"1/2").%%#(\label{lst:dice_meta:pr2}#)
obs((roll(d1),1),true).%%#(\label{lst:dice_meta:obs}#)

h((roll(D),X)) :- roll(D,X).%%#(\label{lst:dice_meta:meta1}#)
roll(D,X) :- h((roll(D),X)).%%#(\label{lst:dice_meta:meta2}#)

\end{lstlisting}

\begin{lstlisting}[float,language=clingos,label={lst:meta},caption={Meta encoding for the frontend of \plog.}]
{ h(A) : random(E,A) } = 1 :- random(E,_).%%#(\label{lst:meta:random}#)

:- not h(A), obs(A, true).%%#(\label{lst:meta:obs1}#)
:-     h(A), obs(A,false).%%#(\label{lst:meta:obs2}#)

h(A) :- do(A).%%#(\label{lst:meta:do}#)

      :~ h((A,V)), not do((A,_)),     pr(E,(A,V),P). [@f1(P),E] %%#(\label{lst:meta:pr}#)
df(E) :- h((A,V)), not do((A,_)), not pr(E,(A,V),_), random(E,(A,_)).%%#(\label{lst:meta:df}#)

:~ df(E), Y = #sum{ @int(P),A : random(E,A),     pr(E,A,P) }.%%#(\label{lst:meta:num1}#)
   [@f2(Y),num,E]%%#(\label{lst:meta:num2}#)
:~ df(E), M = #sum{       1,A : random(E,A), not pr(E,A,_) }.%%#(\label{lst:meta:den1}#)
   [@f3(M),den,E]%%#(\label{lst:meta:den2}#)
\end{lstlisting}

\Splingo\ translates \plog\ programs
combining the translation from \cite{letawa17a} to \lpmln\
with the \transthreezero{} translation from Section~\ref{subsec:lpmlntoplingo}.
Given the input file \texttt{dice.plp},
\splingo\ copies the normal rules
of Lines~\ref{lst:dice:sorts:begin}-\ref{lst:dice:sorts:end},
translates the rules specific to \plog\ into the Listing~\ref{lst:dice_meta},
stores internally the information about the \texttt{\&query} atom, and
adds the general meta-encoding of Listing~\ref{lst:meta}.
In Listing~\ref{lst:dice_meta}, Line~\ref{lst:dice_meta:random}
defines for every dice \texttt{D} one random experiment,
identified by the term \texttt{roll(D)},
that may select for the attribute \texttt{roll(D)} one possible score \texttt{X}.
The atoms defined that way are fed to the first rule of the meta-encoding 
to choose exactly one of those assignments,
represented in this case by an special predicate \texttt{h/1} (standing for "holds"),
that is made equivalent to the predicate \texttt{roll/2}
in Lines~\ref{lst:dice_meta:meta1}-\ref{lst:dice_meta:meta2}
of Listing~\ref{lst:meta}.
Those lines are the interface between the specific input program
and the general meta-encoding.
They allow the latter to refer to the atoms of the former using the predicate
\texttt{h/1}.
Next, Line~\ref{lst:dice_meta:pr2} of Listing~\ref{lst:dice_meta}
defines the probability of the second dice rolling a \texttt{6}
in the experiment identified by the term \texttt{roll(d2)}.
This is used in Line~\ref{lst:meta:pr} of the meta-encoding, 
where \texttt{@f1(P)} returns the logarithm of \texttt{P},
to add that weight whenever the following conditions hold:
the attribute \texttt{A} has the value \texttt{V},
this value has not been fixed by a deliberate action,
and some probabilistic atom gives the probability \texttt{P}.
If there is no such probabilistic atom,
then the rule of Line~\ref{lst:meta:df}
derives that the assignment chosen in the experiment \texttt{E}
receives the default probability,
calculated in Lines~\ref{lst:meta:num1}-\ref{lst:meta:den2}
following the principle of indifference mentioned above,
where \texttt{@f2(Y)} returns the logarithm of \texttt{1-Y},
and \texttt{@f3(M)} returns the logarithm of \texttt{1/M}.
The idea of this calculation is as follows.
For some experiment \texttt{E},
the number \texttt{Y} accounts for the sum of the probabilities of the probabilistic atoms related to \texttt{E},
and \texttt{M} is the number of outcomes of the experiment \texttt{E}
for which there are no probabilistic atoms.
Then,
%
the probability of each outcome of the experiment \texttt{E}
for which there is no probabilistic atom is \texttt{(1-Y)*(1/M)}.
Instead of multiplying those probabilities \texttt{1-Y} and \texttt{1/M},
the encoding adds their logarithms, 
and it does so in two steps:
one in each of the last two weak constraints.
Finally, the observation fact generated in Line~\ref{lst:dice_meta:obs} of Listing~\ref{lst:dice_meta}
is handled by Lines~\ref{lst:meta:obs1}-\ref{lst:meta:obs2} of Listing~\ref{lst:meta},
and the possible deliberate actions, represented by atoms of the form \texttt{do(A)},
are handled in Line~\ref{lst:meta:do} of the meta-encoding.


\section{The system \splingo}\label{sec:system}

\begin{figure}[!h]
    \includegraphics[width=\textwidth]{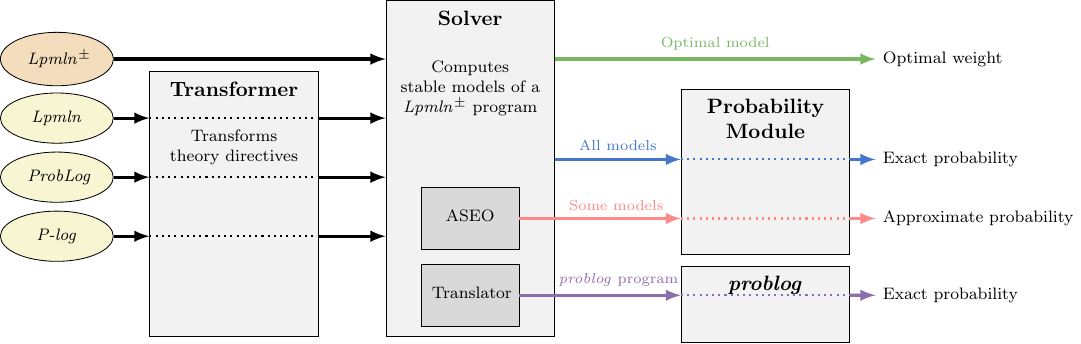}
    \caption{System architecture of \splingo. The frontends are colored in yellow.
        Modules of the system are gray boxes.
        The green flow corresponds to MPE inference,
        the blue one to exact marginal inference,
        and the red one to approximate inference,
        all of them using \splingo's internal solving algorithms.
        The purple flow corresponds to MPE and marginal inference
        using \sproblog.}
    \label{fig:system}
\end{figure}

The implementation of \splingo\ is based on \sclingo\ and its Python API (v5.5,~\citeN{gekakaosscwa16a}).
The system architecture is described in Figure~\ref{fig:system}.
The input is a logic program written in some probabilistic language: \plingo, \lpmln, \problog\ or \plog.
For \plingo, the input language (orange element of Figure~\ref{fig:system})
is the same as the input language of \sclingo,
except for the fact that the weights of the weak constraints can be strings representing real numbers.
For the other languages, the system uses the corresponding frontends,
that translate the input logic programs (yellow elements of Figure~\ref{fig:system})
to the input language of \splingo\ using the Transformer module,
as illustrated by the examples of Section~\ref{sec:plingo}.
Among other things, this involves converting the theory atoms (preceeded by `\texttt{\&}') to normal atoms.
The only exception to this are \texttt{\&query} atoms, that are eliminated from the program
and stored internally.
For \plog, the frontend also appends the meta encoding (Listing~\ref{lst:meta})
to the translation of the input program.

\Splingo\ can be used to solve two reasoning tasks: most probable explanation (MPE) inference and marginal inference.
MPE inference corresponds to the task of finding a most probable stable model of a probabilistic logic program.\footnote{%
    In earlier versions of this work we referred to this as \textit{maximum a posteriori} (MAP).
    However, MAP is in fact the setting where you want to find the most probable assignment of some variables while marginalizing over the others.
    We thank Rafael Kiesel for pointing this out.}
Following the approach of \citeN{letawa17a},
this task is reduced
in \splingo\
to finding one optimal stable model of the input program
using \sclingo's built-in optimization methods.
The only changes that have to be made
concern handling the strings that may occur as weights of weak constraints,
and switching the sign of such weights,
since otherwise \sclingo\ would compute a least probable stable model.
Regarding marginal inference, it can be either applied in general, or with respect to a query.
In the first case, the task is to find all stable models and their probabilities.
In the second case, the task is to find the probability of some query atom,
that is undefined if the input program has no stable models,
and otherwise it is the sum of the probabilities of the stable models that contain that atom.
The basic algorithm of \splingo\ for both cases is the same.
First, the system enumerates all optimal stable models of
the input program excluding the weak constraints at level $0$.
Afterwards, those optimal stable models are passed, together with their costs at level $0$,
to the Probability module, that calculates the required probabilities.

In addition to this exact method (represented by the upper blue arrows in Figure~\ref{fig:system}),
\splingo\ implements an approximation method (red arrows in Figure~\ref{fig:system})
based on the approach presented in \cite{pajjan21a}.
The idea is to simplify the solving process by
computing just a subset of the stable models,
and using this smaller set to approximate the actual probabilities.
Formally, in the definitions of $\weightplingo{\Pi}{X}$ and $\probplingo{\Pi}{X}$,
this implies replacing the set \osmplingo{\Pi} by one of its subsets.
In the implementation, the modularity of this change
is reflected by the fact that the Probability module
is agnostic to whether the stable models that it receives as input are all or just some subset of them.
For marginal inference in general,
this smaller subset consists of $k$ stable models with the highest possible probability,
given some positive integer $k$ that is part of the input.
To compute this subset,
the Solver module of \splingo\ uses a new implementation for the task of
answer set enumeration in the order of optimality (ASEO)
presented in~\cite{pajjan21a}.
Given some positive integer $k$,
the implementation first computes the stable models of the smallest cost,
then, among the remaining stable models, computes the ones of the smallest cost,
and so on until $k$ stable models (if they exist) have been computed.\footnote{%
    The implementation for \sclingo\ \emph{in general} is available at
    \url{https://github.com/potassco/clingo/tree/master/examples/clingo/opt-enum}.
    It applies the multi-shot solving functionality offered by the Python API of \clingo.
    The main method of the implementation mantains a logic program that initially is the same as the input logic program.
    At each computation step this method enumerates all optimal models of the current logic program.
    This is implemented by a call to the \emph{solving} procedure of a \clingo{} \emph{Control} object.
    After each computation step,
    the current program is extended by a ground constraint
    that eliminates all the last enumerated optimal models.
    This constraint guarantees that the optimal stable models found in the next computation step
    are the next best stable models in the order of optimality.
}
For marginal inference with respect to a query,
the smaller subset consists of $k$ stable models containing the query of the highest possible probability,
and another $k$ stable models without the query of the highest possible probability.
In this case, the algorithm for ASEO is set to compute $2\times k$ stable models in total.
But once it has computed $k$ stable models that contain the query,
or $k$ stable models that do not contain the query, whichever happens first,
it adds a constraint enforcing that the remaining stable models fall into the opposite case.

\Splingo\ offers a third solving method
that generates a \problog\ program using
the translation \lpmlntoproblogzero,
and afterwards runs a \sproblog\ solver.
This method can be used for both MPE and marginal inference.
In fact, it can leverage all solving modes and options of a \problog\ solver.
In Figure~\ref{fig:system}, it is represented by the purple arrow.
The Translator component takes as input a \plingo\ program,
and uses the meta-programming option \texttt{--output=reify}
of \clingo~\cite{karoscwa21a}
to ground and reify that program into a list of facts.
Then, in our first approach, we just combined those facts
with a \problog\ meta-encoding implementing the translation \lpmlntoproblogzero.
While this approach was very elegant, it turned out to be inefficient,
because \sproblog~2.2 needed too much time to ground the meta-encoding.
Perhaps a more experienced \problog\ programmer
could develop a more performant meta-encoding.\footnote{%
    The meta-encodings are available at \url{https://github.com/potassco/plingo/tree/master/plingo/meta-problog}.}
In our case, we decided to move the grounding efforts to \clingo.
We developed a meta-encoding that implements the translation \lpmlntoproblogzero,
but replaces the \problog's  probabilistic facts of the form $p :: a$
by ASP facts of the form $prob(p,a)$.
This meta-encoding, together with the reified program,
is grounded by \clingo\ using option \texttt{--text},
resulting in a ground normal logic program.
The final \problog\ program is the combination of this program
with a small \problog\ meta-encoding that creates the corresponding
probabilistic facts given the atoms $\mathit{prob}(p,a)$.
This is the approach that is actually implemented in \splingo.
We would like to highlight that,
although in Section~\ref{subsec:plingotoproblog} we have defined the translation \lpmlntoproblogzero\
only for normal logic programs,
the actual implementation in \splingo\
covers the whole non-disjunctive part of the \clingo\ language~\cite{cafageiakakrlemarisc20a}.
Furthermore, the meta-encoding has been optimized
to minimize the number of new atoms that it introduces.
For example, for programs that follow
the \emph{Generate, Define and Test} methodology~\cite{lifschitz02a} of ASP,
the meta-encoding only introduces the atoms $\posa{a}$ of the translation
for those original atoms $a$
that are not generated by the choice rules \emph{and} do occur in soft formulas.

\section{Experiments}\label{sec:experiments}

In this section, we experimentally evaluate the
three solving methods of \splingo\ version~1.1 and compare them
to native implementations of \lpmln, \problog\ and \plog.\footnote{%
    Available, respectively, at \url{https://github.com/azreasoners/lpmln},
    \url{https://github.com/ML-KULeuven/problog}, and
    \url{https://github.com/iensen/plog2.0}.}
For \lpmln, we evaluate the system \slpmln~\cite{letawa17a}
that is the foundation of the basic implementation of \splingo.
For \problog, we consider the \sproblog\ system version 2.2.2~\cite{fibrreshguthjara15a},
that implements various methods for probabilistic reasoning.
In the experiments, we use one of those methods
that is designed specifically to answer probabilistic queries.
It converts the input program to a weighted Boolean formula
and then applies a knowledge compilation method for weighted model counting.
For \plog, we evaluate two implementations,
that we call \splognaive\ and \splogdco~\cite{bal17a}.
While the former works like \splingo\ and \slpmln\ by enumerating stable models,
the latter implements a different algorithm that builds a computation tree
specific to the input query.
All benchmarks were run on an Intel$^\textnormal{\tiny{TM}}$ 
Xeon E5-2650v4 under Debian GNU/Linux 10,
with 24 GB of memory
and a timeout of 1200 seconds per instance.

We have performed three experiments.
In the first one,
our goal is to evaluate the performance of the three solving methods of \splingo,
and compare it to the performance of all the other systems on the same domain.
In particular,
we want to analyze the solving time and the accuracy
of the approximation method for different values of the input parameter $k$,
and we want to compare the \sproblog-based method of \splingo\
with the system \sproblog\ running an original \problog\ encoding.
In the second experiment,
our goal is to compare \splingo\ with the implementations of \plog\ 
on domains that are specific to this language.
Finally, the goal of the third experiment
is to compare \splingo\ and \slpmln\ on the task of MPE inference.
In particular, in this case,
the basic implementation of \splingo\ and the implementation of \slpmln\
are very similar, and boil down to a single call to \sclingo.
Here, we would like to evaluate if in this situation there is any difference in performance between both approaches.

\begin{figure}
    \centering
    \includegraphics[width=.45\textwidth]{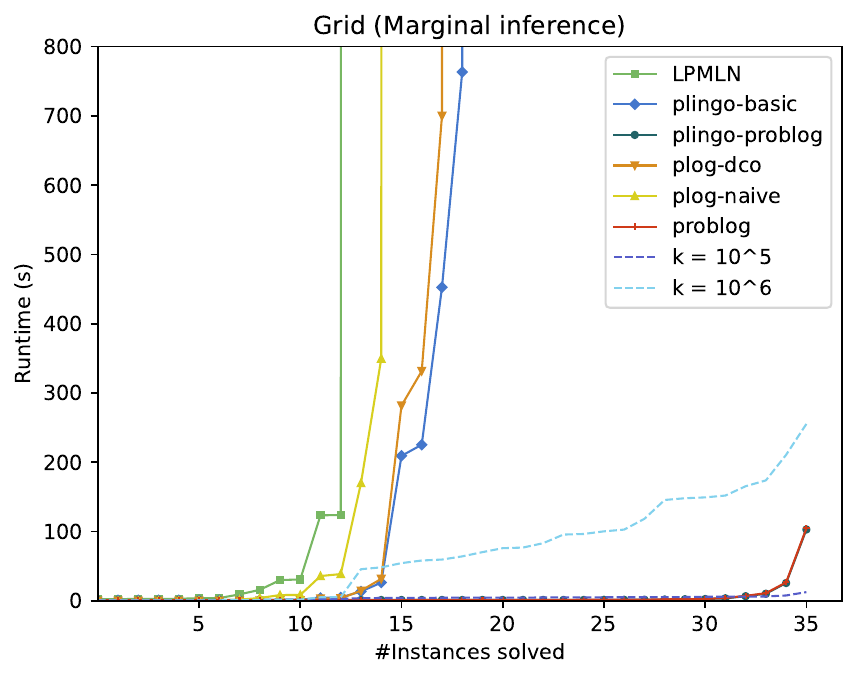}
    \includegraphics[width=.45\textwidth]{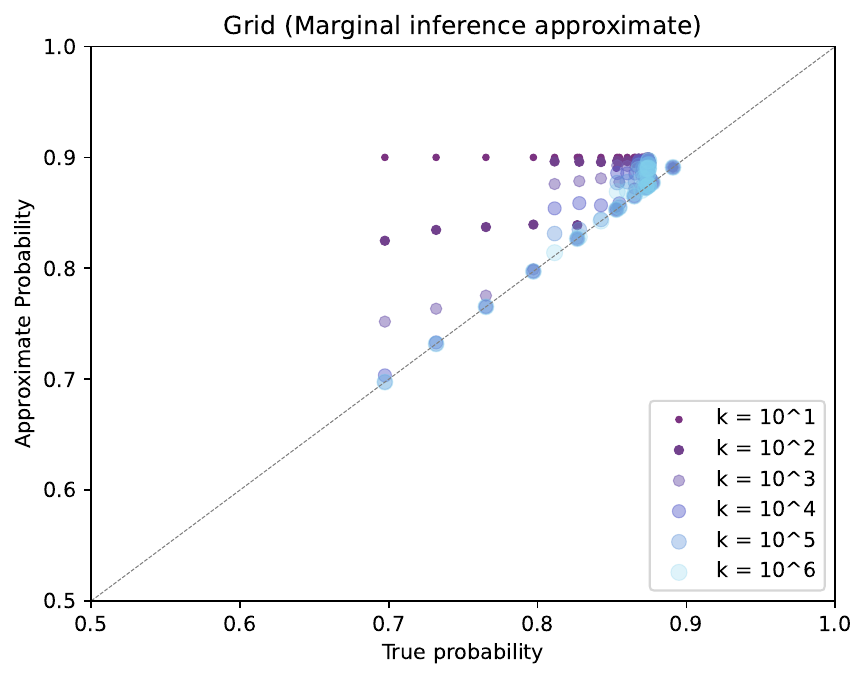}
    \caption{Runtimes of all systems and quality of the approximation method on the Grid domain.}
    \label{fig:results:grid}
\end{figure}

In the first experiment,
we compare all systems on the task of marginal inference with respect to a query
in a probabilistic Grid domain from \cite{zhu12a},
that appeared in a slightly different form in \cite{fibrreshguthjara15a}.
We have chosen this domain because it can be easily and similarly represented
in all these probabilistic languages,
which is required if we want to compare all systems in terms of a single benchmark.
In this domain, there is a grid of size $m \times n$,
where each node $(i,j)$ passes information to the nodes $(i+1,j)$ and $(i,j+1)$ if $(i,j)$ is not faulty,
and each node in the grid can be faulty with probability $0.1$.
The task poses the following question:
what is the probability that node $(m,n)$ receives information from node $(1,1)$?
To answer this, we run exact marginal inference with all approaches,
and approximate marginal inference with \splingo\ for different values of $k$: $10^1$, $10^2$, \ldots, and $10^6$.
The results are shown in Figure~\ref{fig:results:grid}.
On the left side, there is a cactus plot representing how many instances where solved within a given runtime.
The dashed lines represent the runtimes of approximate marginal inference in
\splingo\ for $k=10^5$ and $k=10^6$.
Among the exact implementations,
\sproblog\ and the \sproblog-based method of \splingo\ are the clear winners.
Their solving times are almost the same,
showing that in this case the translation generated by
\splingo\ does not incur in any additional overhead.
The specific algorithm of \sproblog\ for answering queries is much faster
than the other exact systems that either have to enumerate all stable models or,
in the case of \splogdco, may have to explore its whole solution tree.
The runtimes among the rest of the exact systems are comparable,
but \splingo\ is a bit faster than the others.
For the approximation method, on the right side of Figure~\ref{fig:results:grid},
for every value of $k$ and every instance,
there is a dot whose coordinates represent
the probability calculated by the approximation method and the true probability (calculated by \sproblog).
This is complemented by Table~\ref{tab:grid:error},
that shows the average absolute error and the maximal absolute error for each value of $k$ in $\%$,
where the absolute error for some instance and some $k$ in $\%$
is defined as
the absolute value of the difference between the calculated probability and the true probability for that instance,
multiplied by $100$.
We can see that, as the value of $k$ increases,
the performance of the approximation method deteriorates,
while the quality of the approximated probabilities increases.
A good compromise is found for $k=10^5$,
where the runtime is better than \sproblog,
and the average error is below $1\%$.

\begin{table}
\centering
\begin{tabular}{|c||c|c|c|c|c|c|}
\hline
     \textbf{k}           & \textbf{$10^1$} & \textbf{$10^2$} & \textbf{$10^3$} & \textbf{$10^4$} & \textbf{$10^5$} & \textbf{$10^6$} \\
\hline 
              Avg. Error &  $4.7\pm4.3$  & $3.3\pm2.7$   & $2.1\pm1.5$   & $1.4\pm1.2$   & $0.9\pm0.9$ & $0.6\pm0.8$ \\
              Max. Error &  $20.3$          & $12.7$           & $6.5$           & $4.3$           & $2.5$         & $2.3$  \\
\hline
\end{tabular}
\caption{Average and maximal error (in $\%$) of the approximation method on the Grid domain for different values of $k$.}
\label{tab:grid:error}
\end{table}

\begin{figure}
    \includegraphics[width=.45\textwidth]{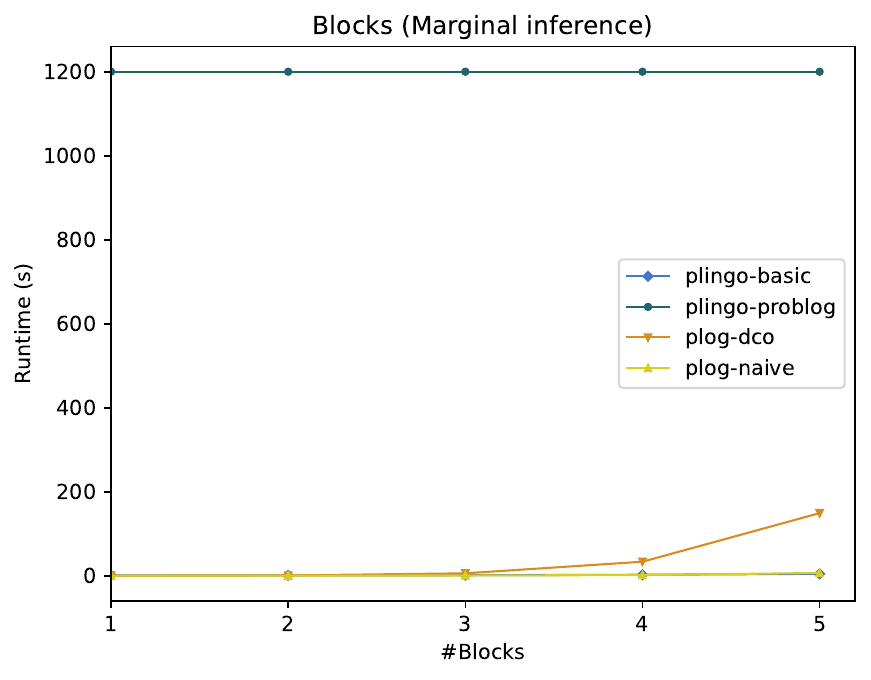}
    \includegraphics[width=.45\textwidth]{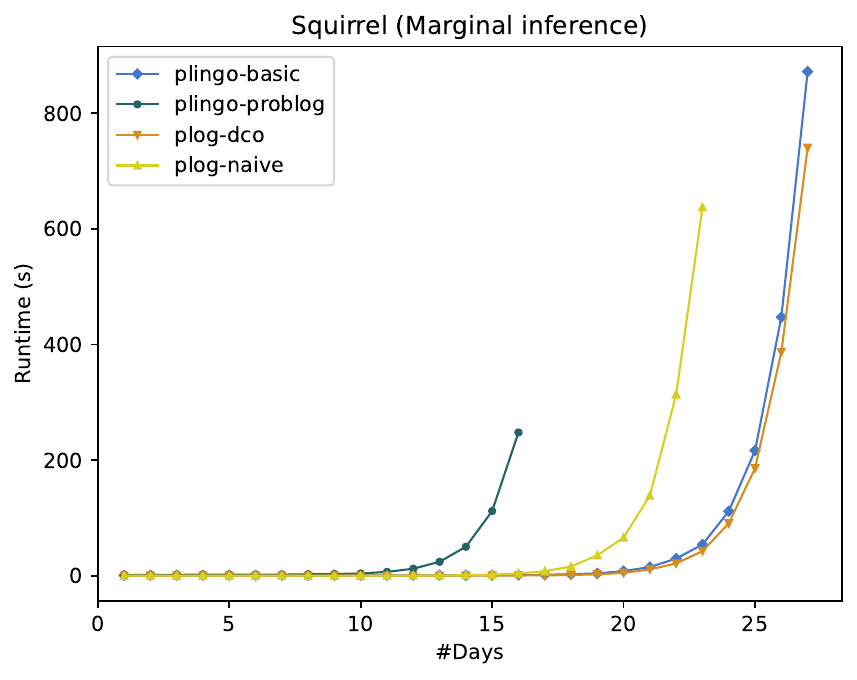}
    \caption{Runtimes of \splingo, \splognaive\ and \splogdco\ on the \plog\ domains.}
    \label{fig:results:plog}
\end{figure}

In the second experiment,
we compare the performance of the two exact methods of \splingo\ using the \plog\ frontend
with the two native implementations of that language, 
on tasks of marginal inference with respect to a query in three different domains: NASA, Blocks and Squirrel.
The NASA domain~\cite{bal17a} involves logical and probabilistic reasoning about the faults of a control system.
For this domain there are only three instances.
All of them are solved by basic \splingo\ in about a second,
while
\sproblog-based \splingo\ takes between $1$ and $80$ seconds,
\splognaive\ between $1$ and $5$ seconds,
and \splogdco\ between $40$ and $100$ seconds.
The Blocks domain~\cite{zhu12a} starts with a grid and a set of $n$ blocks,
and asks what is the probability that two locations are still connected after
the $n$ blocks have been randomly placed on the grid.
In the experiments we use a single map of $20$ locations and vary $n$ between $1$ and $5$.
The results are shown in Figure~\ref{fig:results:plog},
where we can see a similar pattern as in the NASA domain:
basic \splingo\ and \splognaive\ solve all instances in just a few seconds,
while \splogdco\ needs much more time for the bigger instances,
and \splingo\ combined with \sproblog\ times out in all instances.
The Squirrel domain~\cite{bal17a,bageru09a} is an example of Bayesian learning,
where the task is to update the probability of a squirrel finding some food in a patch
on day $n$ after failing to find it on the first day.
In the experiments we vary the number of days $n$ between $1$ and $27$.
The results are shown in Figure~\ref{fig:results:plog}.
In this domain, \sproblog-based \splingo\ can solve instances up to $16$ days,
while \splognaive\ can solve instances up to $23$ days,
and \splingo\ and \splogdco\ can solve instances up to $27$ days.
To interpret the results,
recall that the underlying algorithms of \splognaive\ and \splingo\ are very similar.
Hence, we conjecture that the better performance of \splingo\
is due to details of the implementation.
On the other hand, \splogdco\ uses a completely different algorithm.
According to the authors~\cite{bal17a},
this algorithm should be faster than \splognaive\ when
the value of the query can be determined from a partial assignment of the atoms of the program.
This may be what is happening in the Squirrel domain,
where it is on par with \splingo,
while it does not seem to be the case for the other domains.
We hoped that the \sproblog-based method would allow \splingo\
to solve faster the \plog\ benchmarks,
but the results are in the opposite direction.
In these domains
\clingo\ generates relatively big ground programs in a short time,
and then \problog's solving components take a long time to reason about those programs.
%
%

\begin{figure}
    \includegraphics[width=.45\textwidth]{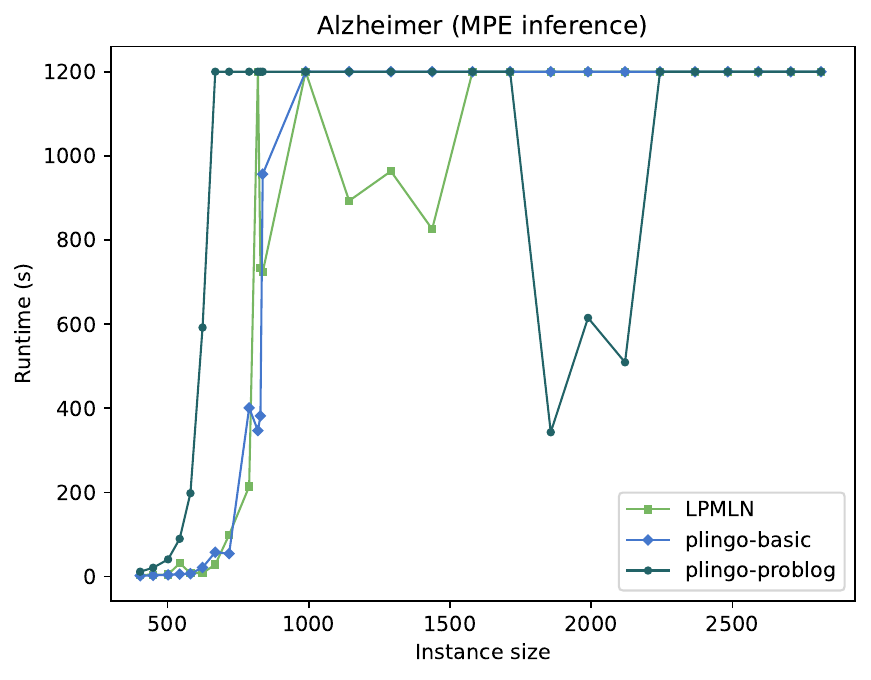}
    \includegraphics[width=.45\textwidth]{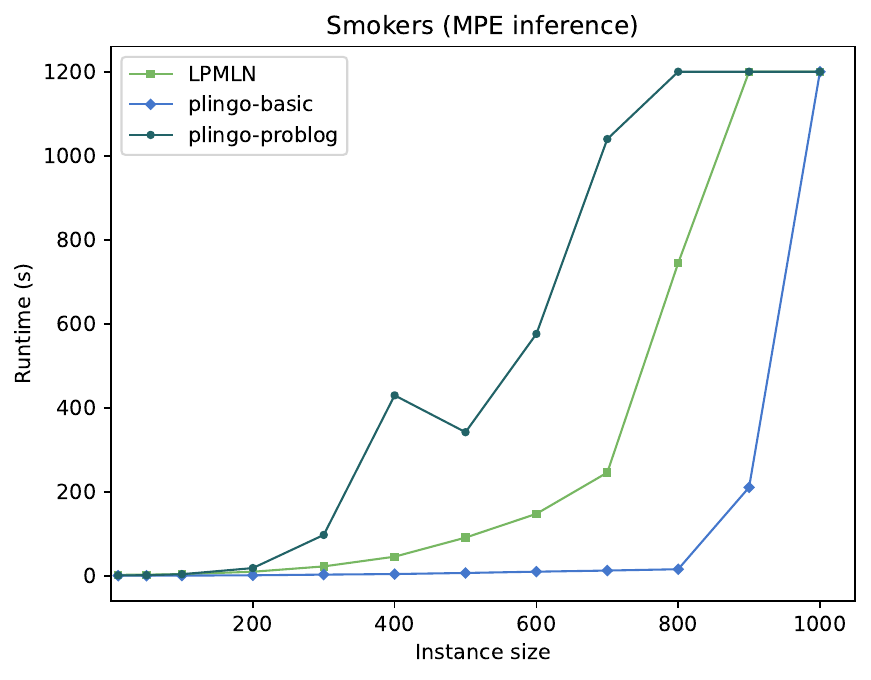}
    \caption{Runtimes of \splingo\ and \slpmln\ on the \lpmln\ domains.}
    \label{fig:results:lpmln}
\end{figure}

In the third experiment,
we compare the performance of the exact methods of \splingo\ using the \lpmln\ frontend
with the system \slpmln\ on tasks of MPE inference in two domains: Alzheimer and Smokers.
The goal in the Alzheimer domain~\cite{rakito07a} is to determine the
probability that two edges are connected in a directed probabilistic graph
based on a real-world biological dataset of Alzheimer genes~\cite{sht15a}.\footnote{%
    We thank Gerda Janssens for providing us the instances.}
The data consists of a directed probabilistic graph with 11530 edges and 5220 nodes.
In the experiments we select different subgraphs of this larger graph,
varying the number of nodes from $100$ to $2800$.
The results are shown in Figure~\ref{fig:results:lpmln},
where we observe that \sproblog-based \splingo\ is slower than the other two methods,
that behave similarly except on some instances of middle size,
that can be solved by \slpmln\ but not by basic \splingo.
The Smokers domain involves probabilistic reasoning about a network of friends.
Originally it was presented in~\cite{rafrkemu08a},
but we use a slightly simplified version from~\cite{letawa17a}.
In the experiments we vary the number of friends in the network.
In Figure~\ref{fig:results:lpmln},
we can observe that basic \splingo\ is the fastest,
followed by \slpmln\ and then by \sproblog-based \splingo.
%
Given that the underlying algorithms of basic \splingo\ and \slpmln\ are similar,
we expected them to have a similar performance.
Looking at the results,
we have no specific explanation for the differences in some instances of the Alzheimer domain,
and we conjecture that they are due to the usual variations in solver performance.
On the Smokers domain, the worse performance of \slpmln\ seems to be due
to the usage of an external parser that increases a lot the preprocessing time for the bigger instances.
With respect to \sproblog-based \splingo,
in both domains it spends most of its time grounding the translation with \clingo,
which seems to explain its worse performace.

\section{Conclusion}\label{sec:conclusion}
\renewcommand{\plingo}{\sysfont{plingo}}

We have presented \plingo, an extension of the ASP system \clingo\ with various probabilistic reasoning modes.
Although based on \lpmln, it also supports \plog\ and \problog.
While the basic syntax of \plingo\ is the same as the one of \clingo,
its semantics relies on re-interpreting
the cost of a stable model at priority level $0$
as a measure of its probability.
Solving exploits the relation between
most probable stable models and optimal stable models~\cite{leeyang17a};
it relies on \clingo's optimization and enumeration modes,
as well as an approximation method based on answer set enumeration in the order of optimality~\cite{pajjan21a}.
This is complemented by another method that is based on a novel translation to \problog.
Our empirical evaluation has shown that \plingo\ is at eye height with other ASP-based probabilistic systems,
and that the different solving methods implemented in the system are complementary.
Notably, the approximation method produced low runtimes and low error rates (below $1\%$) on the Grid domain.
\sysfont{Plingo}\ is freely available at \url{https://github.com/potassco/plingo}.
%

\paragraph{Acknowledgments}
This work was supported by DFG grant SCHA 550/15.
T. Janhunen was partially supported by the Academy of Finland grant 345633 (XAILOG).

\bibliographystyle{acmtrans}

\appendix
\section{Proofs}

\begin{proofof}{Proposition~\ref{prop:ssm}}
    We fix an interpretation $X$.
    First, note that since $\soft{\Pi}$ contains only soft integrity constraints,
    $\unw{\soft{\Pi}}$ contains only (unweighted) integrity constraints.
    Per definition, those integrity constraints which are not satisfied by $X$
    are not included in $\unw{\Pi_X}$.
    The other integrity constraints are satisfied by $X$ and thus do not have any effect on the generation (soft) stable models,
    i.e. $\text{SM}(\unw{\Pi_X}) = \text{SM}(\unw{\Pi_X^{hard}})$.
    Therefore, we have that
    \begin{align*}
        \ssmp{\Pi} & = \{X \mid X \textnormal{ is a (standard) stable model of } \unw{\Pi_X}
        \textnormal{ that satisfies } \unw{\hard{\Pi}} \}                                           \\
                   & = \{X \mid X \textnormal{ is a (standard) stable model of } \unw{\Pi_X^{hard}}
        \textnormal{ that satisfies } \unw{\hard{\Pi}} \}                                           \\
                   & = \ssmp{\Pi^{hard}} \,
    \end{align*}
    Next we prove that
    \begin{enumerate}
        \item $X$ is a stable model of $\unw{\Pi_X^{hard}}$ that satisfies $\unw{\hard{\Pi}}$ iff
        \item $X$ is a stable model of $\unw{\Pi^{hard}}$
    \end{enumerate}
    From the first to the second, if $X$ satisfies $\unw{\hard{\Pi}}$
    then $\unw{\Pi_X^{hard}} = \unw{\Pi^{hard}}$, and so the second follows.
    From the second to the first, if $X$ is a stable model of $\unw{\Pi^{hard}}$
    then it satisfies $\unw{\hard{\Pi}}$ and again $\unw{\Pi_X^{hard}} = \unw{\Pi^{hard}}$
    and the first follows.
    Using this we can simplify the definition of \ssmp{\Pi^{hard}} as follows
    \begin{align*}
        \ssmp{\Pi^{hard}} & = \{ X \mid X \text{ is a (standard) stable model of } \unw{\Pi^{hard}} \} \\
                          & = \text{SM}(\unw{\Pi^{hard}})
    \end{align*}
    This gives us the desired result.

\end{proofof}

\newpage
\begin{lemma}\label{lem:osm_subset}
    Let $\Pi$ be an \lpmln\ program.
    The optimal stable models of $\transtwo{\Pi}$ are a subset of the soft stable models of $\Pi$
    \begin{align*}
        \osm{\transtwo{\Pi}} \subseteq \ssm{\Pi} \, .
    \end{align*}
\end{lemma}
\begin{proofof}{Lemma~\ref{lem:osm_subset}}
    Let $X \in \osm{\transtwo{\Pi}}$.
    Then $X$ is an optimal stable model of
    $ \{ F \vee \neg F \mid w:F \in \Pi \}$,
    and the weak constraints
    $\{ :\sim \, F [-1,1] \mid w : F \in \Pi, w = \alpha \}$.
    So it is a (standard) stable model of $ \{ F \vee \neg F \mid w:F \in \Pi \}$.
    Then $X$ is also a stable model of $\{ F \mid w : F \in \Pi, X \models F \}$.
    But this is just $\unw{\Pi_X}$ and thus $X \in \ssm{\Pi}$.
\end{proofof}

\begin{lemma}\label{lem:translation_alt_probs}
    Let $\Pi$ be an \lpmln\ program.
    If there is at least one interpretation $X$ which satisfies all hard rules in $\Pi$, then
    \begin{align*}
        \probpp{\transthree{\Pi}}{X} = \probpp{\transtwo{\Pi}}{X} \, .
    \end{align*}
\end{lemma}

\begin{proofof}{Lemma~\ref{lem:translation_alt_probs}}
    Note that $\transthree{\Pi}$ and $\transtwo{\Pi}$ differ only in their hard rules
    and that $\transthree{\Pi}$ does not have weak constraints (the hard rules have to be satisfied anyways).
    More specifically the rules in the set $\{  \alpha : F \mid w:F \in \Pi, w = \alpha \}$ in $\transthree{\Pi}$,
    are replaced by a set of choice rules $\{  \alpha : F \vee \neg F \mid w:F \in \Pi, w = \alpha \}$ in $\transtwo{\Pi}$.
    Since there is at least one interpretation $X$ satisfying all hard rules in $\Pi$,
    we know that $\osm{\transthree{\Pi}}$ is not empty. \\
    Further the weak constraints $\{ :\sim \, F [-1,1] \mid w : F \in \Pi, w = \alpha \}$ in $\transtwo{\Pi}$
    guarantee that all optimal stable model of $\transtwo{\Pi}$ satisfy all hard rules.
    Thus $\osm{\transthree{\Pi}} = \osm{\transtwo{\Pi}}$.
    The soft rules in both translations are the same so it follows that
    $\weightpp{\transthree{\Pi}}{X} = \weightpp{\transtwo{\Pi}}{X}$
    and $\probpp{\transthree{\Pi}}{X} = \probpp{\transtwo{\Pi}}{X}$.
\end{proofof}
\par
\medskip
\begin{proofof}{Proposition~\ref{prop:probs_plingo}}
    We first prove the equation for the \textbf{standard semantics}.

    For any interpretation $X$ under the \lpmln\ semantics we have per definition
    \begin{align*}
        \prob{\Pi}{X} = \lim_{\alpha \rightarrow \infty}
        \frac{\weight{\Pi}{X}}{\sum_{Y \in \ssm{\Pi}} \weight{\Pi}{Y}} \, .
    \end{align*}
    Using Lemma~\ref{lem:osm_subset} we can split up the sum in the denominator as follows
    \begin{align*}
        \sum_{Y \in \ssm{\Pi}} \weight{\Pi}{Y} =
        \sum_{Y \in \os{\Pi}} \weight{\Pi}{Y} +
        \sum_{Y \in \nos{\Pi}} \weight{\Pi}{Y} \, ,
    \end{align*}
    where $\os{\Pi} = \osm{\transtwo{\Pi}}$, $\nos{\Pi} = \ssm{\Pi} \setminus \osm{\transtwo{\Pi}}$ and the left sum contains only those interpretations which satisfy some maximal number of hard rules $m$.
    The right sum contains interpretations which satisfy strictly less than $m$ hard rules.
    We can thus also rewrite the weights in the denominator as follows
    \begin{align*}
        \sum_{Y \in \os{\Pi}} \weight{\Pi}{Y} =
        \sum_{Y \in \os{\Pi}} \mathit{exp}(m \alpha) \cdot \tw{\soft{\Pi}_Y} \, ,
    \end{align*}
    and
    \begin{align*}
        \sum_{Y \in \nos{\Pi} } \weight{\Pi}{Y} =
        \sum_{Y \in \nos{\Pi} } \mathit{exp}(n(Y) \cdot \alpha)
        \cdot \tw{\soft{\Pi}_Y} \, ,
    \end{align*}
    with $n(Y) = |\hard{\Pi} \cap \Pi_Y|$.
    \paragraph*{Case 1: $X \in \os{\Pi}$.}
    If $X$ is an optimal stable model of $\transtwo{\Pi}$ it satisfies some maximal number $m$ of hard rules.
    We can rewrite its weight as
    \begin{align*}
        \weight{\Pi}{X} = \mathit{exp}(m \alpha) \cdot \tw{\soft{\Pi}_X} \, .
    \end{align*}
    So the probability becomes
        {\small
            \begin{align*}
                \prob{\Pi}{X} = \lim_{\alpha \rightarrow \infty}
                \frac{\mathit{exp}(m \alpha) \cdot \tw{\soft{\Pi}_X}}
                {\sum_{Y \in \os{\Pi}} \mathit{exp}(m \alpha) \cdot \tw{\soft{\Pi}_Y} +
                \sum_{Y \in \nos{\Pi} } \mathit{exp}(n(Y) \cdot \alpha)
                \cdot \tw{\soft{\Pi}_Y}} \, .
            \end{align*}
        }
    We divide both the numerator and denominator by $\mathit{exp}(m \alpha)$.
        {\small
            \begin{align*}
                \prob{\Pi}{X} = \lim_{\alpha \rightarrow \infty}
                \frac{\tw{\soft{\Pi}_X}}
                {\sum_{Y \in \os{\Pi}} \tw{\soft{\Pi}_Y} +
                \sum_{Y \in \nos{\Pi} }
                \frac{\mathit{exp}(n(Y) \cdot \alpha)}{\mathit{exp}(m \alpha)}
                    \cdot \tw{\soft{\Pi}_Y}} \, .
            \end{align*}
        }
    Since $n(Y) < m$, we conclude that the right sum in the denominator approaches $0$ when $\alpha$ tends to infinity.
    \begin{align*}
        \prob{\Pi}{X} = \frac{\tw{\soft{\Pi}_X}}{\sum_{Y \in \os{\Pi}} \tw{\soft{\Pi}_Y}} \, .
    \end{align*}
    Next, observe that $\soft{\Pi}_X$ contains all soft rules $w : F$ satisfied by $X$.
    If $X$ satisfies $w : F$, then it also satisfies $w : \neg \neg F$ and vice versa;
    therefore, $\soft{\Pi}_X = \soft{\transtwo{\Pi}_X}$ and
    $\tw{\soft{\Pi}_X} = \tw{\soft{\transtwo{\Pi}_X}}$.
    This means we can replace the corresponding expressions in the above equation and obtain the desired result.
    \begin{align*}
        \prob{\Pi}{X} & = \frac{\tw{\soft{\Pi}_X}}{\sum_{Y \in \os{\Pi}} \tw{\soft{\Pi}_Y}}                       \\[0.8em]
                      & = \frac{\tw{\soft{\transtwo{\Pi}_X}}}{\sum_{Y \in \os{\Pi}} \tw{\soft{\transtwo{\Pi}_X}}}
        = \probpp{\transtwo{\Pi}}{X} \, .
    \end{align*}
    \paragraph*{Case 2: $X \in \nos{\Pi}$.}
    If $X$ is not an optimal stable model of $\transtwo{\Pi}$ but it is a soft stable model of $\Pi$,
    per definition its weight and thus its probability are zero under \plingo\ semantics.
    Under \lpmln\ semantics its probability is similar as above,
    only we need to replace $m$ in the numerator by $n(X) = |\hard{\Pi} \cap \Pi_X|$
    {\small
            \begin{align*}
                \prob{\Pi}{X} & = \lim_{\alpha \rightarrow \infty}
                \frac{\mathit{exp}(n(X) \alpha) \cdot \tw{\soft{\Pi}_X}}
                {\sum_{Y \in \os{\Pi}} \mathit{exp}(m \alpha) \cdot \tw{\soft{\Pi}_Y} +
                \sum_{Y \in \nos{\Pi}} \mathit{exp}(n(Y) \cdot \alpha)
                \cdot \tw{\soft{\Pi}_Y}}                              \\[0.8em]
                              & \leq \lim_{\alpha \rightarrow \infty}
                \frac{\mathit{exp}(n(X) \cdot \alpha)}{\mathit{exp}(m \alpha)} \cdot
                \frac{\tw{\soft{\Pi}_X}}{\sum_{Y \in \os{\Pi}} \tw{\soft{\Pi}_Y}} \, .
            \end{align*}
        }
    We can remove the second sum in the denominator since we know that it is always greater or equal than 0.
    Now similar as above we know that $n(X) < m$, so the whole expression approaches $0$.
    It follows
    \begin{align*}
        \prob{\Pi}{X}
        \leq \lim_{\alpha \rightarrow \infty}
        \frac{\mathit{exp}(n(X) \cdot \alpha)}{\mathit{exp}(m \alpha)} \cdot
        \frac{\tw{\soft{\Pi}_X}}{\sum_{Y \in \os{\Pi}} \tw{\soft{\Pi}_Y}}
        = 0 = \probpp{\transtwo{\Pi}}{X} \, .
    \end{align*}
    Since we know that $\prob{\Pi}{X} \geq 0$, we have the desired result.
    \paragraph*{Case 3: $X \notin \ssm{\Pi}$.}
    If $X$ is not a soft stable model, it is neither an optimal stable model.
    Per definition the weight of $X$ is zero under both semantics and thus $\prob{\Pi}{X} = \probpp{\transtwo{\Pi}}{X}$. \\

    \renewcommand{\transtwo}[1]{\ensuremath{\mathit{standard}(#1)}}
    Next we prove the equation for the \textbf{alternative semantics}.
    If there is no interpretation $X$ which satisfies all hard rules in $\Pi$, then $\probp{\Pi}{X}$ is not defined.
    In that case $\osm{\transthree{\Pi}}$ is also empty, since
    \begin{align*}
        \{  \alpha : F \mid w:F \in \Pi, w = \alpha \} \subseteq \unw{\hard{\transthree{\Pi}}} \, .
    \end{align*}
    Thus $\probpp{\transthree{\Pi}}{X}$ is also not defined.
    If $\ssmp{\Pi}$ is not empty,
    then it holds that $\probp{\Pi}{X} = \prob{\Pi}{X}$ for every interpretation $X$
    (cf.~Proposition 2 from \cite{leewan16a}).
    Combining this with Lemma~\ref{lem:translation_alt_probs} gives the desired result
    \begin{align*}
        \probp{\Pi}{X} = \prob{\Pi}{X} = \probpp{\transtwo{\Pi}}{X} = \probpp{\transthree{\Pi}}{X} \, .
    \end{align*}

\end{proofof}

\begin{proofof}{Proposition~\ref{prop:probs_plingo_neg}}
    The proof goes analogously to the proofs of Corollary 1 from \cite{leeyang17a} and Theorem 1 from \cite{letawa17a}. \\
    First of all we show that $\osm{\Pi} = \osm{\transfour{\Pi}}$.
    It is easy to see that $\sm{\Pi} = \sm{\transfour{\Pi}}$.
    It remains to prove that optimality is preserved as well.
    Let $\totalcost{\Pi}{l}$ be the total weak constraint cost at priority level $l$ for program $\Pi$
    and $\cost{\Pi}{X}{l}$ be the cost of interpretation $X$ at priority level $l$.
    If we negate the formulas and flip the weights in the weak constraints (as proposed in the proposition) it holds that
    \begin{align*}
        \cost{\transfour{\Pi}}{X}{l} = \cost{\Pi}{X}{l} - \totalcost{\Pi}{l} \, .
    \end{align*}
    This means that if $\cost{\Pi}{X}{l}$ is minimal in $\Pi$ for some level $l$,
    $\cost{\transfour{\Pi}}{X}{l}$ is minimal in $\transfour{\Pi}$ at that level
    and so optimality is preserved. \\
    Second, we show that $\weightpp{\Pi}{X} = \tw{\soft{\Pi}} \cdot \weightpp{\transfour{\Pi}}{X}$.
    When $X$ is not an optimal stable model of $\Pi$ this is obvious.
    When $X \in \osm{\Pi}$, we have
    \begin{align*}
        \weightpp{\Pi}{X}
         & = \tw{\soft{\Pi_X}}                                                                                                         \\
         & = \mathit{exp} ( \sum_{w:F \in \soft{\Pi} \text{ and } X \models F} w )                                                     \\
         & = \mathit{exp} ( \sum_{w:F \in \soft{\Pi}} w  - \sum_{w:F \in \soft{\Pi} \text{ and } X \nvDash F} w)                       \\
         & = \mathit{exp} ( \sum_{w:F \in \soft{\Pi}} w ) \cdot \mathit{exp} ( - \sum_{w:F \in \soft{\Pi} \text{ and } X \nvDash F} w) \\
         & = \tw{\soft{\Pi}} \cdot \mathit{exp} (\sum_{w:F \in \soft{\transfour{\Pi}} \text{ and } X \models F} w )                    \\
         & = \tw{\soft{\Pi}} \cdot \tw{\soft{\transfour{\Pi}_X}}                                                                       \\
         & = \tw{\soft{\Pi}} \cdot \weightpp{\transfour{\Pi}}{X} \, .
    \end{align*}
    Combining the two results above
    \begingroup
    \addtolength{\jot}{0.8em}
    \begin{align*}
        \probpp{\Pi}{X}
         & = \frac{\weightpp{\Pi}{X}}{\sum_{Y \in \osm{\Pi}} \weightpp{\Pi}{Y}}                                     \\
         & = \frac{\weightpp{\Pi}{X}}{\sum_{Y \in \osm{\transfour{\Pi}}} \weightpp{\Pi}{Y}}                         \\
         & = \frac{\tw{\soft{\Pi}} \cdot \weightpp{\transfour{\Pi}}{X}}
        {\sum_{Y \in \osm{\transfour{\Pi}}} \tw{\soft{\Pi}} \cdot \weightpp{\transfour{\Pi}}{Y}}                    \\
         & = \frac{\weightpp{\transfour{\Pi}}{X}}{\sum_{Y \in \osm{\transfour{\Pi}}} \weightpp{\transfour{\Pi}}{Y}}
        \cdot \frac{\tw{\soft{\Pi}}}{\tw{\soft{\Pi}}}                                                               \\
         & = \frac{\weightpp{\transfour{\Pi}}{X}}{\sum_{Y \in \osm{\transfour{\Pi}}} \weightpp{\transfour{\Pi}}{Y}} \\
         & = \probpp{\transfour{\Pi}}{X} \, .
    \end{align*}
    \endgroup
\end{proofof}
\par
\medskip


\begin{proofof}{Proposition~\ref{prop:problogtoplingo}}
    We consider first the basic case where
    \problogevidence{\Pi} is empty. 

    \emph{Basic case: \problogevidence{\Pi} is empty.}

    Let $\pione$ be the \plingo\ program:
    \begin{align*}
        \problognormal{\Pi}                                                  & \ \cup \
        \{ a \vee \neg a \mid p :: a \in \problogfacts{\Pi}\} \ \cup
        \\
        \{ \mathit{ln}(p) : \neg \neg a \mid p :: a \in \problogfacts{\Pi}\} & \ \cup \
        \{ \mathit{ln}(1-p) : \neg a \mid p :: a \in \problogfacts{\Pi}\}.
    \end{align*}
    By Theorem~4 from~\cite{leewan16a}, and
    Proposition~\ref{prop:probs_plingo} from this paper,
    for every interpretation $X$, it holds that
    $\prob{\Pi}{X}$ and $\probpp{\pione}{X}$ are the same.
    We prove next that $\probpp{\pione}{X}$ and $\probpp{\problogtolpmln{\Pi}}{X}$ are the same.
    This allows us to infer that
    $\prob{\Pi}{X}$ and $\probpp{\problogtolpmln{\Pi}}{X}$ are the same
    and prove this basic case.

    Take any atom $a$ such that $p :: a \in \problogfacts{\Pi}$.
    Let $\pitwo$ be the \plingo\ program
    that results from replacing in $\pione$ the pair of formulas
    $\mathit{ln}(p) : \neg \neg a$ and
    $\mathit{ln}(1-p) : \neg a$ by the single formula $\mathit{ln}(p/(1-p)) : \neg \neg a$.
    We show next that the probabilities defined by
    $\pione$ and $\pitwo$ are the same.
    Since applying this replacement in a sequence to all atoms occurring in probabilistic facts
    we obtain our program $\problogtolpmln{\Pi}$,
    this gives us that $\probpp{\pione}{X}$ and
    $\probpp{\problogtolpmln{\Pi}}{X}$ are the same.

    We show that for all interpretations $X$ it holds that
    \begin{align}\label{eq:problog:weights}
        \weightpp{\pitwo}{X} =
        \weightpp{\pione}{X} / (1-p).
    \end{align}
    Given that $\pione$ and $\pitwo$ only differ in their soft formulas,
    it holds that $\osm{\pione}$ and $\osm{\pitwo}$ are the same.
    Then, if $X \not\in \osm{\pione}$,
    we have that $X \not\in \osm{\pitwo}$, therefore
    $\weightpp{\pione}{X} = \weightpp{\pitwo}{X} = 0$,
    and
    equation~\eqref{eq:problog:weights} holds.
    Otherwise, if $X \in \osm{\pione}$, we consider two cases.
    In the first case, $a \in X$.
    Since the weight of $\neg \neg a$ is $\mathit{ln}(p)$,
    we can represent the value of
    $\weightpp{\pione}{X}$ as
    $\mathit{exp}(\mathit{ln}(p) + \alpha)$
    for some integer $\alpha$.
    Furthermore, given that $\pitwo$ only replaces in $\pione$ the soft formulas for the atom $a$,
    we can also represent the value
    $\weightpp{\pitwo}{X}$ as 
    $\mathit{exp}(\mathit{ln}(p/(1-p)) + \alpha)$.
    Then, equation~\eqref{eq:problog:weights} follows from the next equalities:
    \begin{align*}
        \weightpp{\pitwo}{X} & =
        \mathit{exp}(\mathit{ln}(p/(1-p)) + \alpha)                                             \\
                             & = \mathit{exp}(\mathit{ln}(p/(1-p))) \cdot  \mathit{exp}(\alpha) \\
                             & = \mathit{exp}(\mathit{ln}(p))/(1-p) \cdot  \mathit{exp}(\alpha) \\
                             & = \mathit{exp}(\mathit{ln}(p)+ \alpha)/(1-p)                     \\
                             & = \weightpp{\pione}{X}/(1-p).
    \end{align*}
    In the second case, where $a \notin X$,
    we can represent $\weightpp{\pione}{X}$ as
    $\mathit{exp}(\mathit{ln}(1-p) + \beta)$, and
    $\weightpp{\pitwo}{X}$ as $\mathit{exp}(\beta)$
    for some integer $\beta$.
    Then, equation~\eqref{eq:problog:weights} follows:
    \begin{align*}
        \weightpp{\pitwo}{X} & =
        \mathit{exp}(\beta)                                                                              \\
                             & = \mathit{exp}(\mathit{ln}(1-p) + \beta - \mathit{ln}(1-p))               \\
                             & = \mathit{exp}(\mathit{ln}(1-p) + \beta) / \mathit{exp}(\mathit{ln}(1-p)) \\
                             & = \mathit{exp}(\mathit{ln}(1-p) + \beta) / (1-p)                          \\
                             & = \weightpp{\pione}{X}/(1-p).
    \end{align*}

    With this, we finish the proof for the case where
    $\problogevidence{\Pi}$ is empty
    by showing that the probabilities defined by
    $\pione$ and $\pitwo$ are the same.
    Take any interpretation $X$, and recall that $\osm{\pione}$ and $\osm{\pitwo}$ are the same.
    If $\osm{\pione}$ is empty, then clearly
    \probpp{\pione}{X} and \probpp{\pione}{X} are undefined.
    Otherwise, we have that:
    \begingroup
    \addtolength{\jot}{0.8em}
    \begin{align*}
        \probpp{\pitwo}{X} & = \frac{\weightpp{\pitwo}{X}}{\sum_{Y \in \osm{\pitwo}} \weightpp{\pitwo}{Y}}                       \\
                           & = \frac{\weightpp{\pione}{X}/(1-p)}{\sum_{Y \in \osm{\pione}} \big(\weightpp{\pione}{Y}/(1-p)\big)} \\
                           & = \frac{\weightpp{\pione}{X}/(1-p)}{\big(\sum_{Y \in \osm{\pione}} \weightpp{\pione}{Y}\big)/(1-p)} \\
                           & = \frac{\weightpp{\pione}{X}}{\sum_{Y \in \osm{\pione}} \weightpp{\pione}{Y}}                       \\
                           & = \probpp{\pione}{X}
    \end{align*}
    \endgroup
    \newpage
    \emph{General case.}

    Given a \problog\ program $\Pi$,
    by $\problogmodels{\Pi}$ we denote
    the set of interpretations $X$ such that $\prob{\Pi}{X}$ is greater than $0$.
    We show that
    \begin{align}\label{eq:problog:general}
        \osm{\pifinal} = \problogmodels{\Pi}.
    \end{align}
    We consider first the program without evidence $\pifinalproblog$.
    Take any interpretation $X$, and consider two cases.

    In the first case, $X$ belongs to \osm{\pifinalnoev}.
    The probability \probpp{\pifinalnoev}{X} is a fraction whose
    numerator is the exponential of some number
    and whose denominator is greater or equal than the numerator.
    Hence, $\probpp{\pifinalnoev}{X} > 0$.
    Given the basic case that we proved above,
    this implies that $\prob{\pifinalproblog}{X} > 0$ and
    therefore $X$ belongs to $\problogmodels{\pifinalproblog}$.

    In the second case, $X$ does not belong to \osm{\pifinalnoev}.
    Hence, $\probpp{\pifinalnoev}{X}$ is $0$ or undefined.
    Given the basic case that we proved above,
    this implies that $\prob{\pifinalproblog}{X}$ cannot be greater than $0$
    and therefore $X$ does not belong to $\problogmodels{\pifinalproblog}$.

    Both cases together prove equation~\eqref{eq:problog:general} for program $\pifinalproblog$.
    The result for program $\Pi$ in general follows from this,
    given that
    \begin{itemize}
        \item
              ${\osm{\pifinalnoev}=\problogmodels{\pifinalproblog}}$,
        \item
              \osm{\pifinal} is the subset of \osm{\pifinalnoev} that satisfies all literals in $\problogevidence{\Pi}$, and
        \item
              \problogmodels{\Pi} is the subset of \problogmodels{\pifinalproblog}
              that satisfies all literals in $\problogevidence{\Pi}$.
    \end{itemize}
    Once equation~\eqref{eq:problog:general} is proved,
    we can approach the general case of this proposition.

    We consider first the case where \osm{\pifinal} is empty.
    On the one side, this implies that
    \probpp{\pifinal}{X} is undefined for every interpretation $X$.
    On the other side,
    \problogmodels{\Pi} is also empty by equation~\eqref{eq:problog:general},
    and therefore
    \prob{\Pi}{X} is also undefined for every interpretation $X$.

    Now, we consider the case where \osm{\pifinal} is not empty.
    Note that this implies that the probabilities of $\Pi$ and $\pifinal$ are all defined.
    Take any interpretation $X$.
    If $\probpp{\pifinal}{X}$ is $0$ then $X$ does not belong to \osm{\pifinal}.
    Hence, $X$ does not belong to $\problogmodels{\Pi}$ by equation~\eqref{eq:problog:general},
    and $\prob{\Pi}{X}$ is also $0$.
    We prove the case where $\probpp{\pifinal}{X}$ is greater than $0$
    by the equalities below.
    Note that this case implies that $X$ belongs to $\osm{\pifinal}$.
    By $\ssum{\Pi}$ we denote the sum
    \[\sum_{Y \in \osm{\pifinalnoev}} \weightpp{\pifinalnoev}{Y}.\]
    \newpage
    \begingroup
    \addtolength{\jot}{0.8em}
    \begin{align*}
        \probpp{\pifinal}{X} & = \frac{\weightpp{\pifinal}{X}}{\sum_{Y \in \osm{\pifinal}} \weightpp{\pifinal}{Y}}       \\
                             & = \frac{\weightpp{\pifinal}{X} / \ssum{\Pi}}
        {\big(\ \sum_{Y \in \osm{\pifinal}} \weightpp{\pifinal}{Y}\big) / \ssum{\Pi}}                                    \\
                             & = \frac{\weightpp{\pifinal}{X} / \ssum{\Pi}}
        {\sum_{Y \in \osm{\pifinal}} \big(\weightpp{\pifinal}{Y} / \ssum{\Pi}\big)}                                      \\
                             & = \frac{\weightpp{\pifinalnoev}{X} / \ssum{\Pi}}
        {\sum_{Y \in \osm{\pifinal}} \big(\weightpp{\pifinalnoev}{Y} / \ssum{\Pi}\big)}                                  \\
                             & = \frac{\probpp{\pifinalnoev}{X}}{\sum_{Y \in \osm{\pifinal}} \probpp{\pifinalnoev}{Y}}   \\
                             & = \frac{\prob{\pifinalproblog}{X}}{\sum_{Y \in \osm{\pifinal}} \prob{\pifinalproblog}{Y}} \\
                             & = \frac{\probbasic{\Pi}{X}}{\sum_{Y \in \osm{\pifinal}} \probbasic{\Pi}{Y}}               \\
                             & = \frac{\probbasic{\Pi}{X}}{\sum_{Y \in \problogmodels{\Pi}} \probbasic{\Pi}{Y}}          \\
                             & = \prob{\Pi}{X}
    \end{align*}
    \endgroup
    The first equality holds by definition of $\probpp{\pifinal}{X}$.
    The second is the result of dividing both sides of the fraction by $\ssum{\Pi}$.
    The third rearranges the formulas in the denominator.
    The fourth replaces
    \weightpp{\pifinal}{X} by
    \weightpp{\pifinalnoev}{X}
    and
    \weightpp{\pifinal}{Y} by
    \weightpp{\pifinalnoev}{Y}.
    This replacement is sound given that $X$ and the $Y$'s belong to \osm{\pifinal},
    and in this case the substraction of $\problogevidence{\Pi}$
    does not affect their corresponding weights.
    The fifth equation holds by definition of $\probpp{\pifinalnoev}{X}$.
    The sixth holds by the basic case that we proved above.
    The seventh holds by definition of $\probbasic{\Pi}{X}$.
    The eigth holds by equation~\eqref{eq:problog:general}.
    Finally, the ninth equation holds by definition of $\prob{\Pi}{X}$.
\end{proofof}

\par
\medskip
\begin{proofof}{Proposition~\ref{prop:plingotoproblog}}
    %
    %
    We show that the probabilities remain the same
    through the steps of the translations of Section~\ref{subsec:plingotoproblog}.
    We start at step 2.
    Given a set $Y$, by $Y'$ we denote its extension with copy atoms $Y \cup \{\posa{a}\mid a \in Y\}$.
    %

    \emph{Step 2.}
    Let $\Gamma$ be the union of $\{\neg \mybot\}$
    with the set of rules~\eqref{eq:copy} for every atom $a \in \atoms{\Pi}$,
    and $\Pi_2$ be $\Pi \cup \Gamma$ (we jump program $\Pi_1$).
    We focus now on the hard part of $\Pi_2$.
    By the Splitting Set Theorem for propositional formulas~\cite{ferraris11a},
    $\hard{\Pi_2}$ can be splitted into $\hard{\Pi}$ and $\Gamma$.
    Then, $Y$ is a stable model of $\hard{\Pi_2}$ iff
    $Y$ is a stable model of $\Gamma\cup Z$
    for some stable model $Z$ of $\hard{\Pi}$.
    The stable models $Y$ of $\Gamma\cup Z$ are exactly
    the sets $Z\cup Z'$.
    To see this, observe that:
    \begin{itemize}
        \item
              The set of atoms occurring in the heads of the rules of $\Gamma\cup Z$
              is $Z\cup\{\posa{a}\mid a \in \atoms{\Pi}\}$.\footnote{%
                  For this purpose, we see the formula $\neg \mybot$ as the rule without head $\bot \leftarrow \mybot$.}
              Hence,
              $Y \subseteq  Z\cup\{\posa{a}\mid a \in \atoms{\Pi}\}$.
        \item
              The formulas in $\Gamma\cup Z$ classically entail the facts $Z$,
              and they also entail that $a \equiv \posa{a}$ for every atom $a \in \atoms{\Pi}$.
              This, together with the previous item, implies that $Y$ must have the form
              $Z\cup Z'$.
        \item
              Every atom in
              $Z\cup Z'$
              is justified, either by a fact in $Z$, or by a choice rule in $\Gamma$.
    \end{itemize}
    Given the previous statements,
    it follows that the set of stable models of \hard{\Pi_2}
    is
    $\{Z \cup Z' \mid Z \textnormal{ is a stable model of }\hard{\Pi}\}$.
    Since \weak{\Pi} and \weak{\Pi_2} are empty,
    the sets \osm{\Pi} and \osm{\Pi_2}
    consist of
    the sets of stable models of $\hard{\Pi}$ and $\hard{\Pi_2}$,
    respectively.
    Then, it follows that the set
    \osm{\Pi_2} is
    $\{Z \cup Z' \mid Z \in \osm{\Pi}\}$.
    Finally,
    since the soft formulas of $\Pi_2$ are the same as those of $\Pi$,
    and they refer only to the atoms in \atoms{\Pi},
    we can conclude that
    for every interpretation $X$, disjoint from $\{\posa{a}\mid a \in \atoms{\Pi}\}$,
    it holds that
    $\weightpp{\Pi_2}{X \cup X'} = \weightpp{\Pi}{X}$
    and $\probpp{\Pi_2}{X \cup X'} = \probpp{\Pi}{X}$.

    \emph{Step 3.}
    Let $\tr{\Pi}$ be the set of rules $\{\tr{r}\mid r \in \hard{\Pi}\}$
    and $\Pi_3$ be the program
    $(\Pi_2 \setminus \hard{\Pi}) \cup \tr{\Pi}$.
    Program $\Pi_3$ can also be represented as
    $\tr{\Pi}\cup \Gamma \cup \soft{\Pi}$.
    We show below that the stable models of $\hard{\Pi_2}$ and $\hard{\Pi_3}$ are the same.
    Then, given that the soft formulas of both programs do not change,
    we can conclude that
    $\weightpp{\Pi_3}{X \cup X'} = \weightpp{\Pi_2}{X \cup X'} = \weightpp{\Pi}{X}$
    and
    $  \probpp{\Pi_3}{X \cup X'} =   \probpp{\Pi_2}{X \cup X'} =   \probpp{\Pi}{X}$
    for every interpretation $X$ disjoint from $\{\posa{a}\mid a \in \atoms{\Pi}\}$.

    For similar reasons as in Step 2,
    the stable models of $\hard{\Pi_3}$ must have the form $Z \cup Z'$
    for some set $Z \subseteq \atoms{\Pi}$.
    The same holds for the stable models of $\hard{\Pi_2}$.
    Hence, we can consider only interpretations of that form.
    We say that those interpretations are \emph{valid}.
    We prove that the stable models of $\hard{\Pi_2}$ and $\hard{\Pi_3}$
    are the same by showing that for every valid interpretation $X$
    the reduct of $\hard{\Pi_2}$ with respect to $X$ is the same program as
    the reduct of $\hard{\Pi_3}$ with respect to $X$.
    Recall that the reduct of a program with respect to an interpretation is
    the result of replacing in that program every maximal subformula
    that is not satisfied by $X$ by $\bot$.
    The rules in $\Gamma$ are the same in $\hard{\Pi_2}$ and $\hard{\Pi_3}$,
    hence their reduct is also the same in both programs.
    Then, we only have to consider the rules $r \in \hard{\Pi}$ of the form:
    \[
        a_0 \leftarrow a_1 \wedge \ldots \wedge a_m \wedge \neg {a_{m+1}}, \ldots, \neg {a_n}
    \]
    and their translation $\tr{r} \in \tr{\Pi}$:
    \[
        a_0 \leftarrow a_1 \wedge \ldots \wedge a_m \wedge \neg {\posa{a}_{m+1}}, \ldots, \neg {\posa{a}_n}.
    \]
    Let $B(r)$ denote the body of $r$, and $B(\tr{r})$ denote the body of $\tr{r}$.
    Note that since $X$ is valid,
    it satisfies any literal in $r$ iff it satisfies the corresponding literal in $\tr{r}$.
    This also implies that $X$ satisfies $B(r)$ iff $X$ satisfies $B(\tr{r})$.
    We consider four cases and show that for each of them the reducts of $r$ and $\tr{r}$ are the same:
    \begin{itemize}
        \item
              $X$ neither satisfies $a_0$ nor $B(r)$:
              then $X$ does not satisfy $B(\tr{r})$,
              and the reduct of both $r$ and $\tr{r}$ wrt.\ $X$ is
              $\bot \leftarrow \bot$;
        \item
              $X$ does not satisfy $a_0$ but it satisfies $B(r)$:
              then $X$ satisfies $B(\tr{r})$,
              and the reduct of both $r$ and $\tr{r}$ wrt.\ $X$ is
              $\bot \leftarrow a_1 \wedge \ldots \wedge a_m$;
        \item
              $X$ satisfies $a_0$ but does not satisfy $B(r)$:
              then $X$ does not satisfy $B(\tr{r})$,
              and the reduct of both $r$ and $\tr{r}$ wrt.\ $X$ is
              $a_0 \leftarrow \bot$;
        \item
              $X$ satisfies $a_0$ and $B(r)$:
              then $X$ satisfies $B(\tr{r})$,
              and the reduct of both $r$ and $\tr{r}$ wrt.\ $X$ is
              $a_0 \leftarrow a_1 \wedge \ldots \wedge a_m$.
    \end{itemize}

    \emph{Step 4.}
    Let $\Pi_4$ the union of $\hard{\Pi_3}$ and the following soft formulas:
    \begin{align*}
        \{ w : \neg \neg \posa{a} \mid w : \neg \neg a \in \soft{\Pi} \} \ \cup \
        \{ 0 : \neg \neg \posa{a} \mid a \in \atoms{\Pi}\setminus\softatoms{\Pi}\}
    \end{align*}
    that replace the soft formulas $w : \neg \neg a$ from $\soft{\Pi_3}$.
    The sets \osm{\Pi_3} and \osm{\Pi_4} are the same
    given that $\Pi_3$ and $\Pi_4$ only differ in their soft formulas.
    This implies that if $X$ belongs to $\osm{\Pi_4}$ then $X$ is valid.
    Given such a valid $X$,
    let $\Omega_3$ and $\Omega_4$
    denote the set of soft formulas $\soft{\alpha}_{X}$ for $\alpha=\Pi_3$ and $\alpha=\Pi_4$,
    respectively.
    The following equivalences show that
    $\tw{\Omega_3}$ is the same as $\tw{\Omega_4}$:
    \begin{align*}
        \tw{\Omega_3}
         & = \mathit{exp}(\sum_{\{w : \neg \neg a \in \Pi_3 \ \mid \ X \textnormal{ satisfies }\neg \neg a\}}w)               \\
         & = \mathit{exp}(\sum_{\{w : \neg \neg a \in \Pi_3 \ \mid \ X \textnormal{ satisfies }\neg \neg \posa{a}\}}w)        \\
         & = \mathit{exp}(\sum_{\{w : \neg \neg \posa{a} \in \Pi_4 \ \mid \ X \textnormal{ satisfies }\neg \neg \posa{a}\}}w) \\
         & = \tw{\Omega_4}
    \end{align*}
    The first equality holds by definition of \tw{\Omega_3};
    the second holds given that $X$ is valid;
    the third one holds given that $w : \neg \neg a \in \Pi_3$ implies that $w : \neg \neg \posa{a} \in \Pi_4$,
    and the remaining soft formulas of $\Pi_4$ have weight $0$;
    and the fourth equality holds by definition of \tw{\Omega_4}.

    All in all, we have that the sets \osm{\Pi_3} and \osm{\Pi_4} are the same,
    and for every $X$ that belongs to them it holds that $\tw{\Omega_3}=\tw{\Omega_4}$.
    From this, we conclude that
    $\weightpp{\Pi_4}{X \cup X'} = \weightpp{\Pi_3}{X \cup X'} = \weightpp{\Pi}{X}$
    and
    $  \probpp{\Pi_4}{X \cup X'} =   \probpp{\Pi_3}{X \cup X'} =   \probpp{\Pi}{X}$
    for every interpretation $X$ disjoint from $\{\posa{a}\mid a \in \atoms{\Pi}\}$.

    \emph{Step 5.}
    We consider the \problog\ program \lpmlntoproblog{\Pi},
    that is the result of replacing in $\Pi_4$
    the choice rules
    \[ \{ \posa{a} \vee \neg \posa{a} \mid a \in \atoms{\Pi}\} \]
    and the soft formulas
    \begin{align*}
        \{ w : \neg \neg \posa{a} \mid w : \neg \neg a \in \soft{\Pi} \} \ \cup \
        \{ 0 : \neg \neg \posa{a} \mid a \in \atoms{\Pi}\setminus\softatoms{\Pi}\}
    \end{align*}
    by the probabilistic facts
    \begin{align*}
        \{ \frac{e^w} {e^w + 1} :: \posa{a} \mid w : \neg \neg a \in \soft{\Pi} \} \ \cup \
        \{ 0.5 :: \posa{a} \mid a \in \atoms{\Pi}\setminus\softatoms{\Pi}\}.
    \end{align*}
    In turn, the program \problogtolpmln{\lpmlntoproblog{\Pi}}
    is the result of replacing in \lpmlntoproblog{\Pi} those probabilistic facts
    by the choice rules
    \begin{align*}
        \{ \posa{a} \vee  \neg \posa{a} \mid w : \neg \neg a \in \soft{\Pi} \} \ \cup \
        \{ \posa{a} \vee \neg \posa{a} \mid a \in \atoms{\Pi}\setminus\softatoms{\Pi}\}
    \end{align*}
    and the soft formulas
    \begin{align*}
        \bigg\{
        \mathit{ln}\bigg( \frac{ \frac{e^w}{e^w + 1} }{ 1-\frac{e^w}{e^w + 1}} \bigg)
        :: \posa{a} \mid w : \neg \neg a \in \soft{\Pi} \bigg\}                                                            & \ \cup \\
        \bigg\{ \mathit{ln}\bigg(\frac{0.5}{1-0.5}\bigg) :: \posa{a} \mid a \in \atoms{\Pi}\setminus\softatoms{\Pi}\bigg\} & .
    \end{align*}
    This program~\problogtolpmln{\lpmlntoproblog{\Pi}}
    is the same as $\Pi_4$,
    once we simplify the weights of the soft formulas.
    The choice rules are the same,
    given that the first set of choice rules can also be represented as
    $\{ \posa{a} \vee  \neg \posa{a} \mid a \in \softatoms{\Pi} \}$
    and $\atoms{\Pi}=\softatoms{\Pi}\cup(\atoms{\Pi}\setminus\softatoms{\Pi})$.
    The soft formulas are the same because
    \[
        \mathit{ln}\bigg( \frac{ \frac{e^w}{e^w + 1} }{ 1-\frac{e^w}{e^w + 1}} \bigg)
        =
        \mathit{ln}\bigg( \frac{ \frac{e^w}{e^w + 1} }{ \frac{e^w + 1 - e^w}{e^w + 1}} \bigg)
        =
        \mathit{ln}\bigg( \frac{ \frac{e^w}{e^w + 1} }{ \frac{1}{e^w + 1}} \bigg)
        =
        \mathit{ln}\bigg( \frac{ e^w }{ 1 } \bigg) = w
    \]
    and
    \[
        \mathit{ln}\bigg(\frac{0.5}{1-0.5}\bigg) =
        \mathit{ln}(1) = 0.
    \]
    From Proposition~\ref{prop:problogtoplingo},
    for every interpretation $X$ the probabilities
    \prob{\lpmlntoproblog{\Pi}}{X} and
    \probpp{\problogtolpmln{\lpmlntoproblog{\Pi}}}{X}
    are the same.
    Since
    \problogtolpmln{\lpmlntoproblog{\Pi}}
    is the same as $\Pi_4$,
    the probabilities
    \prob{\lpmlntoproblog{\Pi}}{X} and
    \probpp{\Pi_4}{X}
    are also the same.
    Given this and the results of step 4,
    we can conclude that
    $\prob{\lpmlntoproblog{\Pi}}{X \cup X'} =   \probpp{\Pi_4}{X \cup X'} = \probpp{\Pi}{X}$
    for every interpretation $X$ disjoint from $\{\posa{a}\mid a \in \atoms{\Pi}\}$,
    and finish the proof.
\end{proofof}

\end{document}